\documentclass[nohyperref]{article}

\usepackage{microtype}
\usepackage{graphicx}
\usepackage{subfigure}
\usepackage{booktabs} % for professional tables

\usepackage{breakurl}

\usepackage{hyperref}

\usepackage[accepted]{icml2023}

\usepackage{amsmath}
\usepackage{amssymb}
\usepackage{mathtools}
\usepackage{amsthm}

\usepackage{tikzit}
\tikzstyle{node}=[fill=white, draw=black, shape=circle, minimum size=1mm, ultra thick]
\tikzstyle{small box}=[fill=white, draw=black, shape=rectangle, minimum height=0.5cm, minimum width=0.5cm, ultra thick]
\tikzstyle{weyl}=[fill=white, draw={rgb,255: red,0; green,0; blue,109}, shape=rectangle, minimum height=0.5cm, minimum width=0.5cm]
\tikzstyle{filled_node}=[fill=black, draw=black, shape=circle, minimum size=1mm, ultra thick]

\tikzstyle{thick}=[-, ultra thick]
\tikzstyle{blue_thick}=[-, ultra thick, draw=blue]
\tikzstyle{dashes}=[-, dashed, draw={rgb,255: red,191; green,191; blue,191}, dash pattern=on 2mm off 1mm, fill={rgb,255: red,244; green,228; blue,0}]
\tikzstyle{thick_arrow}=[ultra thick, ->]
\tikzstyle{dash_1}=[-, dashed]
\tikzstyle{dash_2}=[-, dashed, fill={rgb,255: red,246; green,235; blue,255}]
\tikzstyle{dash_3}=[-, dashed, fill={rgb,255: red,229; green,255; blue,181}]
\tikzstyle{dash_4}=[-, dashed, fill={rgb,255: red,255; green,209; blue,153}]
\tikzstyle{red_thick}=[-, ultra thick, draw=red]
\tikzstyle{dash_5}=[-, dashed, fill={rgb,255: red,225; green,255; blue,254}]

\usepackage{bm}
\usepackage{nicematrix}
\usepackage{comment}
\usepackage{tabularray}

\DeclareMathOperator{\End}{End}

\DeclareMathOperator{\Hom}{Hom}

\DeclareMathOperator{\sgn}{sgn}

\usepackage[capitalize,noabbrev]{cleveref}


\theoremstyle{plain}
\newtheorem{theorem}{Theorem}[section]

\theoremstyle{definition}

\theoremstyle{remark}
\newtheorem{remark}[theorem]{Remark}

\newtheorem{defn}[theorem]{Definition}

\usepackage[textsize=tiny]{todonotes}

\icmltitlerunning{
	Brauer's Group Equivariant Neural Networks
}

\begin{document}

\twocolumn[
\icmltitle{
	Brauer's Group Equivariant Neural Networks
}




\begin{icmlauthorlist}
\icmlauthor{Edward Pearce--Crump}{imperial}
%\icmlauthor{Firstname2 Lastname2}{equal,yyy,comp}
%\icmlauthor{Firstname3 Lastname3}{comp}
%\icmlauthor{Firstname4 Lastname4}{sch}
%\icmlauthor{Firstname5 Lastname5}{yyy}
%\icmlauthor{Firstname6 Lastname6}{sch,yyy,comp}
%\icmlauthor{Firstname7 Lastname7}{comp}
%\icmlauthor{Firstname8 Lastname8}{sch}
%\icmlauthor{Firstname8 Lastname8}{yyy,comp}
\end{icmlauthorlist}

\icmlaffiliation{imperial}{Department of Computing, Imperial College London, United Kingdom}
%\icmlaffiliation{comp}{Company Name, Location, Country}
%\icmlaffiliation{sch}{School of ZZZ, Institute of WWW, Location, Country}

\icmlcorrespondingauthor{Edward Pearce--Crump}{ep1011@ic.ac.uk}

\icmlkeywords{Machine Learning, ICML}

\vskip 0.3in
]



\printAffiliationsAndNotice{}  % leave blank if no need to mention equal contribution

\begin{abstract}
	We provide a full characterisation of all of the possible group equivariant neural networks 
	whose layers are some tensor power of $\mathbb{R}^{n}$
	for three symmetry groups
	that are missing from the machine learning literature:
	$O(n)$, the orthogonal group; $SO(n)$, the special orthogonal group; and $Sp(n)$, the symplectic group.
	In particular,	
	we find a spanning set of matrices for the learnable, linear, equivariant layer functions between such tensor power spaces
	in the standard basis of $\mathbb{R}^{n}$ when the group is 
	$O(n)$ or $SO(n)$,
	and in the symplectic basis of $\mathbb{R}^{n}$ when the group is 
	$Sp(n)$.
\end{abstract}

\section{Introduction}

Finding neural network architectures that are equivariant to a symmetry group has been an active area of research ever since it was first shown how convolutional neural networks, which are equivariant to translations, could be used to learn from images.
%, video and audio data.
Unlike with multilayer perceptrons, however, the requirement for the overall network to be equivariant to the symmetry group typically 
restricts the form of the network itself.
%the possible form of these neural network architectures is typically restricted by .
Moreover, since these networks exhibit parameter sharing within each layer, 
%often 
ordinarily far fewer parameters appear in these networks than in multilayer perceptrons.
This 
%often 
usually results in simpler, more interpretable models that generalise better to unseen data.

Symmetry groups naturally appear in problems coming from physics,
%The source of many of these symmetry groups comes from physics, 
where the data 
that 
%arises from 
is 
generated by
a physical process often 
comes with 
%has
a certain type of symmetry 
that is 
baked into the data itself. 
The data is typically high dimensional, and it
can often be
%is often 
represented in the form of a high order tensor 
%product 
so that
%in order to capture 
complex relationships can be captured between different features in the data.
%can be captured.
%\textbf{INSERT SOMETHING ABOUT WHY TENSOR PRODUCTS ARE RELEVANT FOR DATA ANALYSIS HERE?}
Consequently, it is important to be able to construct neural networks that can learn efficiently from such data.

%\textbf{INSERT MY OTHER NEW SENTENCE HERE?}
There are two approaches that are typically used for constructing group equivariant neural networks.
%for many symmetry groups.
The first 
%approach 
%has involved 
employs a universal approximation theorem to learn functions that are approximately equivariant, such as in \cite{kumagai}.
The second 
%approach 
involves
%The typical approach for constructing equivariant neural networks for many symmetry groups has involved 
decomposing tensor product representations of the symmetry group in question into irreducible representations.
For example, neural networks that are equivariant to the special orthogonal group $SO(3)$ \cite{clebschgordan},
the special Euclidean group $\mathit{SE}(3)$ \cite{3dSteerable}, and the proper orthochronous Lorentz group $SO^{+}(1,3)$ \cite{bogatskiy}
all use irreducible decompositions 
%for these groups 
and the resulting change of basis transformations into Fourier space 
in their implementations.
%to implement them.
%these neural networks.
However, for most groups, finding this irreducible decomposition is 
not trivial, since the relevant Clebsch--Gordan coefficients are typically unknown.
%a hard problem, 
Furthermore, even if such a decomposition can be found, the resulting neural networks 
%that are constructed 
are often inefficient since forward and backward Fourier transforms are usually required to perform the calculations, which come with a high computational cost.

In this paper, we take an entirely different approach, one which results in a full characterisation of all of the possible group equivariant neural networks whose layers are some tensor power of $\mathbb{R}^{n}$ for the following three symmetry groups: 
$O(n)$, the orthogonal group; $SO(n)$, the special orthogonal group; and $Sp(n)$, the symplectic group.
%In particular, we find a spanning set of matrices for the learnable, linear, equivariant layer functions between such tensor power spaces in the standard basis of $\mathbb{R}^{n}$ when the group is $O(n)$ or $SO(n)$, and in the symplectic basis of $\mathbb{R}^{n}$ when the group is $Sp(n)$.
%In doing so, we avoid having to calculate any irreducible decompositions for the tensor power spaces, and therefore avoid having to perform any Fourier transforms to change the basis of the tensor power spaces.
Our approach is motivated by a mathematical concept that, to the best of our knowledge, has not appeared in any of the machine learning literature to date, other than in \cite{pearcecrump}.
%where we apply a similar approach to the one seen in this paper to provide a full characterisation of all of the possible permutation equivariant neural networks whose layers are some tensor power space of $\mathbb{R}^{n}$.
In that paper, they use that the symmetric group is in Schur--Weyl duality with the partition algebra to provide a full characterisation of
all of the possible permutation equivariant neural networks whose layers are some tensor power of $\mathbb{R}^{n}$.
In this paper, we use as our motivation the results given in 
%a paper of Brauer's, 
\textit{On Algebras Which are Connected with the Semisimple Continuous Groups}~\cite{Brauer}.
Brauer showed that the orthogonal group is in Schur--Weyl duality with an algebra called the Brauer algebra; that the symplectic group is also in Schur--Weyl duality with the Brauer algebra, and that the special orthogonal group is in Schur--Weyl duality with an algebra which we have termed the Brauer--Grood algebra.
By adapting the combinatorial diagrams that form a basis for these algebras, 
%to form vector spaces instead, 
%and the argument given in Brauer's paper, 
we are able to find a spanning set of matrices for the learnable, linear, equivariant layer functions between tensor power spaces of $\mathbb{R}^{n}$ in the standard basis of $\mathbb{R}^{n}$ when the group is $O(n)$ or $SO(n)$, and in the symplectic basis of $\mathbb{R}^{n}$ when the group is $Sp(n)$.
%This makes the group equivariant neural networks in question simple to implement.
In doing so, we avoid having to calculate any irreducible decompositions for the tensor power spaces, and therefore avoid having to perform any Fourier transforms to change the basis accordingly.
%of the tensor power spaces.

The main contributions of this paper are as follows:
\begin{enumerate}
	%\item We introduce the Brauer and Brauer--Grood vector spaces that each have a basis of diagrams indexed by certain set partitions
		%These diagrams can be used to find a spanning set of matrices for 
		%the learnable, linear, equivariant layer functions 
		%between tensor power spaces of $\mathbb{R}^{n}$
		%in the standard basis of $\mathbb{R}^{n}$ when the group is 
	%$O(n)$ or $SO(n)$,
	%and in the symplectic basis of $\mathbb{R}^{n}$ when the group is 
	%$Sp(n)$.
	\item We are the first to show how the combinatorics underlying the Brauer and Brauer--Grood vector spaces, adapted from the Schur--Weyl dualities established by Brauer \yrcite{Brauer},
		provides the theoretical background for constructing group equivariant neural networks for the orthogonal, special orthogonal, and symplectic groups when the layers are some tensor power of $\mathbb{R}^{n}$.
	\item We find a spanning set of matrices for the learnable, linear, equivariant layer functions between such tensor power spaces in the standard basis of $\mathbb{R}^{n}$ when the group is $O(n)$ or $SO(n)$, and in the symplectic basis of $\mathbb{R}^{n}$ when the group is $Sp(n)$.
	%\item The neural networks that we characterise are simple to implement since we 
		%%Our approach makes it simple to implement these neural networks since we 
		%avoid having to consider the decomposition of the tensor power spaces of $\mathbb{R}^{n}$ into irreducible representations for each of the three groups in question. 
		%%which makes it simple to implement these neural networks.
		%%which is known to be a hard problem in most cases.
	\item We generalise our diagrammatical approach to show how to construct neural networks that are equivariant to local symmetries.
	\item 
	We suggest that Schur--Weyl duality is a
	powerful mathematical concept that 
	%has not appeared in the machine learning literature to date, but we remark that it is an approach that 
	could be used to characterise other group equivariant neural networks beyond those considered in this paper.
		%the three groups in question.
		%neural networks that are equivariant to other groups.
\end{enumerate}

\section{Preliminaries}

%We choose our field of scalars to be $\mathbb{C}$ throughout, although the results generalise to any field $\mathbb{F}$ of characteristic zero, such as $\mathbb{R}$.
%Tensor products are also taken over $\mathbb{C}$, unless otherwise stated.
We choose our field of scalars to be $\mathbb{R}$ throughout. 
Tensor products are also taken over $\mathbb{R}$, unless otherwise stated.
Also, we let $[n]$ represent the set $\{1, \dots, n\}$. 

%Recall that a representation of a group $G$ is a choice of vector space $V$ over $\mathbb{C}$ and a group homomorphism
Recall that a representation of a group $G$ is a choice of vector space $V$ over $\mathbb{R}$ and a group homomorphism
\begin{equation} \label{grouprephom}
	\rho : G \rightarrow GL(V)	
\end{equation}
We choose to focus on finite-dimensional vector spaces $V$ 
that are some tensor power of $\mathbb{R}^{n}$
in this paper.

We often abuse our terminology by calling $V$ a representation of $G$, even though the representation is technically the homomorphism $\rho$.
When the homomorphism $\rho$ needs to be emphasised alongside its vector space $V$, we will use the notation $(V, \rho)$.

\section{Group Equivariant Neural Networks} \label{Groupequivnnssection}

Group equivariant neural networks are constructed by alternately composing linear and non-linear $G$-equivariant maps between representations of a group $G$. 
The following is based on the material presented in \cite{lim}.

We first define \textit{$G$-equivariance}:

%We define \textit{$G$-equivariance} below:
\begin{defn} \label{Gequivariance}
	Suppose that $(V, \rho_{V})$ and $(W, \rho_{W})$
	are two representations of a group $G$.

	A map $\phi : V \rightarrow W$ is said to be $G$-equivariant if,
	for all $g \in G$ and $v \in V$,
\begin{equation} \label{Gequivmapdefn}
	\phi(\rho_{V}(g)[v]) = \rho_{W}(g)[\phi(v)]
	%\text{ for all } g \in G \text{ and } v \in V.
\end{equation}
	The set of all \textit{linear} $G$-equivariant maps between $V$ and $W$ is denoted by $\Hom_{G}(V,W)$. 
	When $V = W$, we write this set as $\End_{G}(V)$.
	It can be shown that $\Hom_{G}(V,W)$ is a vector space over $\mathbb{R}$, and that $\End_{G}(V)$ is an algebra over $\mathbb{R}$.
	See \cite{segal} for more details. 
\end{defn}

A special case of $G$-equivariance is \textit{$G$-invariance}:
\begin{defn}
	The map $\phi$ given in Definition \ref{Gequivariance} is said to be $G$-invariant if $\rho_{W}$ is defined to be the $1$-dimensional trivial representation of $G$.
	As a result, $W = \mathbb{R}$.
	%\textbf{CHECK THIS!}
\end{defn}

%\begin{remark}
	%Since the $d$--dimensional trivial representation of $O(n)$ can be decomposed into a direct sum of $d$ irreducible representations of $O(n)$, each of 
	%which is the irreducible $1$--dimensional trivial representation of $O(n)$, we choose in the following to focus on $O(n)$--invariant functions that map to the $1$--dimensional trivial representation $\mathbb{R}$. \textbf{CHECK THIS!}
%\end{remark}

We can now define the type of neural network that is the focus of this paper:
\begin{defn} \label{Gneuralnetwork}
	An $L$-layer $G$-equivariant neural network $f_{\mathit{NN}}$ is a composition of \textit{layer functions}
	\begin{equation}
		f_{\mathit{NN}} \coloneqq f_L \circ \ldots \circ f_{l} \circ \ldots \circ f_1
  	\end{equation}
	such that the $l^{\text{th}}$ layer function is a map of representations of $G$
	\begin{equation}
		f_l: (V_{l-1}, \rho_{l-1}) \rightarrow (V_l, \rho_l)
	\end{equation}
	that is itself a composition
  	\begin{equation} \label{Glayerfidefn}
	  	f_l \coloneqq \sigma_l \circ \phi_l
  	\end{equation}
	of a learnable, linear, $G$-equivariant function $\phi_l : (V_{l-1}, \rho_{l-1}) \rightarrow (V_l, \rho_l)$ together with 
	a fixed, non-linear activation function $\sigma_l: (V_l, \rho_l) \rightarrow (V_l, \rho_l)$ 
	such that
	\begin{enumerate}
		\item $\sigma_l$ is a $G$-equivariant map, as in (\ref{Gequivmapdefn}), and
		\item $\sigma_l$ acts pointwise (after a basis has been chosen for each copy of $V_l$ in $\sigma_l$.)
	\end{enumerate}
\end{defn}

We focus on the learnable, linear, $G$-equivariant functions in this paper because the non-linear functions are fixed. 

\begin{comment}
\begin{figure}[ht]
	\begin{center}
	\scalebox{0.5}{\input{Gequivariantnns.tikz}}
	\caption{Group Equivariant Neural Networks.}
  	\label{Gequivariantnns}
	\end{center}
\end{figure}
\end{comment}

\begin{remark}
	The entire neural network $f_{\mathit{NN}}$ is itself a $G$-equivariant function because 
	it can be shown that
	the composition of any number of $G$-equivariant functions is itself $G$-equivariant.
\end{remark}

\begin{remark}
	One way of making 
	a neural network of the form given in Definition \ref{Gneuralnetwork}
	$G$-invariant 
	is by choosing the representation in the final layer
	to be the $1$-dimensional trivial representation of $G$. 
\end{remark}

\begin{comment}
\begin{remark}
	Figure \ref{Gequivariantnns} highlights the equivariance property of an $G$-equivariant neural network.

	Specifically, the figure shows that if some $v_0 \in V_0$ is passed through the neural network (as represented by the left hand dotted box) with output equal to some $v_L \coloneqq f_{\mathit{NN}}(v_0) \in V_L$, then if $\rho_0(g)[v_0]$, for any $g \in G$, is passed through the same neural network (as represented by the right hand dotted box), the output will be 
	\begin{equation}	
		\rho_L(g)[v_L] = \rho_L(g)[f_{\mathit{NN}}(v_0)] = f_{\mathit{NN}}(\rho_0(g)[v_0])
	\end{equation}
	that is, the neural network respects the action of the group $G$ on both its input and its output.

	It is important to emphasise that each dotted box in Figure \ref{Gequivariantnns} represents the same neural network $f_{\mathit{NN}}$, that is, the network architecture itself does not change between the dotted boxes.
\end{remark}
\end{comment}

\section{The groups $G = O(n)$, $SO(n)$, and $Sp(n)$} \label{groupsOnSOnSpn}

We consider throughout the real vector space $\mathbb{R}^{n}$.
%of dimension $n$, for some $n \in \mathbb{Z}_{\geq 1}$.

Let $GL(n)$ be the group of invertible linear transformations from $\mathbb{R}^{n}$ to $\mathbb{R}^{n}$.
If we pick a basis for each copy of $\mathbb{R}^{n}$, 
then for each linear map in $GL(n)$ we obtain its matrix representation in the bases of $\mathbb{R}^{n}$ that were chosen.
Let $SL(n)$ be the subgroup of $GL(n)$ consisting of all invertible linear transformations from $\mathbb{R}^{n}$ to $\mathbb{R}^{n}$ whose determinant is $+1$.

We can associate to $\mathbb{R}^{n}$ one of the following two bilinear forms:
\begin{enumerate}
	\item a non-degenerate, symmetric bilinear form 
$(\cdot{,}\cdot): \mathbb{R}^{n} \times \mathbb{R}^{n} \rightarrow \mathbb{R}$.
	\item a non-degenerate, skew-symmetric bilinear form $\langle\cdot{,}\cdot\rangle : \mathbb{R}^{n} \times \mathbb{R}^{n} \rightarrow \mathbb{R}$. 
		In this case, $n$ must be even, say $n = 2m$, as a result of applying Jacobi's Theorem. 
		See page 6 of Goodman and Wallach~\yrcite{goodman} for more details. 
\end{enumerate}

Then we can define the groups $O(n)$, $SO(n)$, and $Sp(n)$ as follows:
\begin{enumerate}
	%\item $O(n) \coloneqq \{ g \in GL(n) \mid (gu,gv) = (u,v) \text{ for all } \\ u, v \in \mathbb{R}^{n} \}$
	\item $O(n) \coloneqq \left\{ g \in GL(n) \;\middle\vert
		\begin{array}{l}
		(gx,gy) = (x,y) \\
		\text{ for all } x, y \in \mathbb{R}^{n} 
		\end{array}
		\right\}$
	\item $SO(n) \coloneqq O(n) \cap SL(n)$
	%\item For $n = 2m$, $Sp(n) \coloneqq \{ g \in GL(n) \mid \langle gx,gy \rangle = \langle x,y \rangle \text{ for all } x, y \in \mathbb{R}^{n} \}$.
	\item $Sp(n) \coloneqq \left\{ g \in GL(n) \;\middle\vert
		\begin{array}{l}
		\langle gx,gy \rangle = \langle x,y \rangle \\
		\text{ for all } x, y \in \mathbb{R}^{n} 
		\end{array}
		\right\}$,
		for $n = 2m$.
\end{enumerate}
It can be shown that each of these groups are subgroups of $GL(n)$.

There are special bases of $\mathbb{R}^{n}$ with respect to each of the forms given above.

Firstly, for the form $(\cdot{,}\cdot)$, by Lemma 1.1.2 on page 4 of Goodman and Wallach~\yrcite{goodman},
we may assume that there is an ordered basis 
\begin{equation}
B \coloneqq \{e_1, e_2, \dots, e_n\}
\end{equation}
of $\mathbb{R}^{n}$,
where $e_i$ has a $1$ in the $i^{\text{th}}$ position, and a $0$ elsewhere,
which satisfies the relations
\begin{equation} \label{orthonormalbasis}
	(e_i, e_j) = \delta_{i,j}
\end{equation}
with respect to the form
$(\cdot{,}\cdot)$. 
The basis $B$ is called the \textit{standard basis} for $\mathbb{R}^{n}$, and by (\ref{orthonormalbasis}),
it is an orthonormal basis of $\mathbb{R}^{n}$.
It is clear that the matrix representation of the form 
$(\cdot{,}\cdot)$
in the basis $B$ is the $n \times n$ identity matrix.

Hence the form
$(\cdot{,}\cdot)$
in the basis $B$ is the Euclidean inner product
\begin{equation} \label{euclideaninnprod}
	(x,y) = x^\top y \text{ for all } x, y \in \mathbb{R}^{n}
\end{equation}
where 
$x$ is the column vector 
$(x_1, x_{2}, \dots, x_n)^\top$ 
and
$y$ is the column vector 
$(y_1, y_{2}, \dots, y_n)^\top$ 
when expressed in the basis $B$.

Secondly, for the form $\langle\cdot{,}\cdot\rangle$, where $n = 2m$, by Lemma 1.1.5 on page 7 of Goodman and Wallach~\yrcite{goodman},
we may assume that there is an ordered basis 
\begin{equation}
	\widetilde{B} \coloneqq \{e_1, e_{1'}, \dots, e_m, e_{m'}\}
\end{equation}
of $\mathbb{R}^{n}$, where the $i^{\text{th}}$ basis vector in the set has a $1$ in the $i^{\text{th}}$ position and a $0$ elsewhere, 
which satisfies the relations
\begin{equation}
	\langle e_\alpha, e_\beta \rangle = \langle e_{\alpha'}, e_{\beta'} \rangle = 0
\end{equation}
%and
\begin{equation}
	\langle e_\alpha, e_{\beta'} \rangle = - \langle e_{\alpha'}, e_{\beta} \rangle = \delta_{\alpha, \beta}
\end{equation}
with respect to the form
$\langle\cdot{,}\cdot\rangle$. 
The basis $\widetilde{B}$ is called the \textit{symplectic basis} for $\mathbb{R}^{n}$. 

\begin{comment}
\textbf{POTENTIALLY CAN CUT THIS UP TO HENCE}
Indeed, by the same Lemma, picking a basis of $\mathbb{R}^{n}$ that satisfies a number of relations with respect to the form
$\langle\cdot{,}\cdot\rangle$
is equivalent to first giving an $n \times n$ non-singular, skew-symmetric matrix $\Omega$
that we choose to be the matrix representation of the form
$\langle\cdot{,}\cdot\rangle$
in some basis of $\mathbb{R}^{n}$ and then working out the basis itself.

In this case, $\Omega$ is the block diagonal matrix
\begin{equation}
	\NiceMatrixOptions{xdots/shorten = 0.6 em}
	\begin{pNiceMatrix}
		\bm{J} & \bm{0} & \Cdots & \bm{0} \\
		\bm{0} & \bm{J} & \Ddots & \Vdots \\
		\Vdots & \Ddots & \bm{J} & \bm{0} \\
		\bm{0} & \Cdots & \bm{0} & \bm{J}
		\CodeAfter \line{2-2}{3-3}
	\end{pNiceMatrix}
\end{equation}
where $\bm{J}$ is the $2 \times 2$ matrix
\begin{equation}
	\begin{pmatrix}
		0 & -1 \\
		1 & \phantom{-}0 
	\end{pmatrix}
\end{equation}
and $\bm{0}$ is the
$2 \times 2$ zero matrix.
\end{comment}

Hence the form
$\langle\cdot{,}\cdot\rangle$
in the basis $\widetilde{B}$ is
the skew product
%can be expressed as
\begin{equation} \label{skewsymmform}
	\langle x,y \rangle 
	=
	\sum_{r = 1}^{m} \left(x_r y_{r'} - x_{r'} y_{r}\right)
	=
	\sum_{i,j} \epsilon_{i,j} x_i y_{j}
\end{equation}
for all $x, y \in \mathbb{R}^{n}$,
where 
$x$ is the column vector 
$(x_1, x_{1'}, \dots, x_m, x_{m'})^\top$ 
expressed in the basis $\widetilde{B}$,
$y$ is the column vector 
$(y_1, y_{1'}, \dots, y_m, y_{m'})^\top$ 
expressed in the basis $\widetilde{B}$,
and
%\begin{equation}
	%x = x_1v_1 + x_{1'}v_{1'} + \dots + x_mv_{m} + x_{m'}v_{m'}
%\end{equation}
%\begin{equation}
	%y = y_1v_1 + y_{1'}v_{1'} + \dots + y_mv_{m} + y_{m'}y_{m'}
%\end{equation}
\begin{equation} \label{epsilondef1}
	\epsilon_{\alpha, \beta} = \epsilon_{{\alpha'}, {\beta'}} = 0
\end{equation}
\begin{equation} \label{epsilondef2}
	\epsilon_{\alpha, {\beta'}} = - \epsilon_{{\alpha'}, {\beta}} = \delta_{\alpha, \beta}
\end{equation}
%Then it can easily be shown, using (\textbf{INSERT REF}),
%that $g \in Sp(n)$ if and only if $A^\top \Omega A = \Omega$, where $A$ is the $n \times n$ matrix representation of $g$ in the basis $B$,
%and $\Omega$ is as given in (\textbf{INSERT REF}).

\section{The space $(\mathbb{R}^{n})^{\otimes k}$ as a representation of $G$}

Let $G$ be any of the groups $O(n)$, $SO(n)$, and $Sp(n)$ (where $n=2m$ for $Sp(n)$).

Then, for any positive integer $k$, the space
$(\mathbb{R}^{n})^{\otimes k}$ 
is a representation of $G$, denoted by $\rho_k$, where
\begin{equation}
	\rho_k(g)(v_1 \otimes \dots \otimes v_k)
	\coloneqq
	gv_1 \otimes \dots \otimes gv_k
\end{equation}
for all $g \in G$ and for all vectors $v_i \in \mathbb{R}^{n}$. 
We call 
$(\mathbb{R}^{n})^{\otimes k}$ 
the $k$-order tensor power space
of $\mathbb{R}^{n}$.

Moreover, 
each form 
%on $\mathbb{R}^{n}$
$\mathbb{R}^{n} \times \mathbb{R}^{n} \rightarrow \mathbb{R}$
induces a~non-degenerate bilinear form 
%on $(\mathbb{R}^{n})^{\otimes k}$, 
$(\mathbb{R}^{n})^{\otimes k} \times (\mathbb{R}^{n})^{\otimes k} \rightarrow \mathbb{R}$,
given by
\begin{equation}	
	(v_1 \otimes \dots \otimes v_k, w_1 \otimes \dots \otimes w_k)
	\coloneqq
	\prod_{r = 1}^{k} (v_r, w_r)
\end{equation}
for the symmetric case, and
\begin{equation}	
	\langle v_1 \otimes \dots \otimes v_k, w_1 \otimes \dots \otimes w_k\rangle
	\coloneqq
	\prod_{r = 1}^{k} \langle v_r, w_r \rangle
\end{equation}
for the skew-symmetric case.

Consequently, there is a standard basis for 
$(\mathbb{R}^{n})^{\otimes k}$ 
that is
induced from the standard basis for  
$\mathbb{R}^{n}$
for the symmetric case,
and, similarly, there is a symplectic basis for
$(\mathbb{R}^{n})^{\otimes k}$ 
that is
induced from the symplectic basis for  
$\mathbb{R}^{n}$
for the skew-symmetric case.

Our goal is to characterise all of the possible learnable, linear $G$-equivariant layer functions between any two tensor power spaces of $\mathbb{R}^{n}$.
In doing so, we will be able to characterise and implement all of the possible $G$-equivariant neural networks whose layers are a tensor power space of $\mathbb{R}^{n}$.

Specifically, we want to find, ideally, a basis, or, at the very least, a spanning set, of matrices for 
%a $k$--order to $l$--order $G$--equivariant layer function
%$(\mathbb{R}^{n})^{\otimes k} \rightarrow (\mathbb{R}^{n})^{\otimes l}$ 
$\Hom_G((\mathbb{R}^{n})^{\otimes k}, (\mathbb{R}^{n})^{\otimes l})$
when either the standard basis (for $G = O(n), SO(n)$) or the
symplectic basis (for $G = Sp(n), n = 2m$) is chosen for $\mathbb{R}^{n}$. 

Note that the $G$-invariance case is encapsulated within this, since this 
%is just the case where 
occurs when $l = 0$.

\begin{remark} \label{remarkfeatureone}
	We assume throughout most of the paper that the feature dimension for all of our representations is one.
	This is because 
	the group $G$ does not act on the feature space.
	We relax this assumption in Section~\ref{featuresbiases}.

	Also, the layer functions under consideration do not take into account any bias terms, but we will show in Section~\ref{featuresbiases}
	that these can be easily introduced.
\end{remark}

\section{
	A Spanning Set for 
	$\Hom_G((\mathbb{R}^{n})^{\otimes k}, (\mathbb{R}^{n})^{\otimes l})$
	%, the space of learnable, linear, $G$--equivariant functions
} \label{SpanningSetResults}

Brauer's~\yrcite{Brauer} paper \textit{On Algebras Which are Connected with the Semisimple Continuous Groups} focused, in part, on calculating a spanning set of matrices for 
$\End_G((\mathbb{R}^{n})^{\otimes k})$
in the standard basis of $\mathbb{R}^{n}$
for the groups $G = O(n)$ and $SO(n)$, and in the symplectic basis of $\mathbb{R}^{n}$ for $Sp(n)$. 
%when $n = 2m$.
%where $G$ is either $O(n)$, $SO(n)$, or (for $n=2m$) $Sp(n)$,
%in the standard basis (for $G = O(n), SO(n)$) or the symplectic basis (for $G = Sp(n), n = 2m$) of $\mathbb{R}^{n}$.

Brauer achieved this by applying the First Fundamental Theorem of Invariant Theory for each of the groups in question (see the Technical Appendix) to a spanning set of invariants, one for each group $G$, that he showed are in bijective correspondence with the spanning set of matrices for 
$\End_G((\mathbb{R}^{n})^{\otimes k})$.

In particular, for each group $G$, he associated a diagram
with each element of the spanning set of invariants,
which we describe in further detail below.
In doing so,
he constructed a bijective correspondence between a set of such diagrams and a spanning set of matrices for 
$\End_G((\mathbb{R}^{n})^{\otimes k})$
in the standard/symplectic basis of $\mathbb{R}^{n}$.

We want to find, instead, a spanning set of matrices for 
$\Hom_G((\mathbb{R}^{n})^{\otimes k}, (\mathbb{R}^{n})^{\otimes l})$
in the standard/symplectic basis of $\mathbb{R}^{n}$.
The results that we describe in the following will contain, as a special case, Brauer's results, since when $l = k$, we see that
\begin{equation}
\Hom_G((\mathbb{R}^{n})^{\otimes k}, (\mathbb{R}^{n})^{\otimes l})
= 
\End_G((\mathbb{R}^{n})^{\otimes k})
\end{equation}

Brauer considered two types of diagrams 
coming from certain set partitions,
whose definitions we adapt for our purposes.
The first is defined as follows:
\begin{defn}
	For any $k, l \in \mathbb{Z}_{\geq 0}$, a $(k,l)$--Brauer partition $\beta$ is a partitioning of the set $[l+k]$ 
	%\coloneqq \{1, \dots, l+k\}$ 
	into a disjoint union of pairs.
	We call each pair a block.
	%Each pair is called a block, which are all of size $2$.
	Clearly, if $l+k$ is odd, then no such partitions exist.
	Therefore, in the following, we assume that $l+k$ is even.
	By convention, if $k = l = 0$, then there is just one $(0,0)$--Brauer partition.

%For 
	We can represent 
	each $(k,l)$--Brauer partition $\beta$
	%it 
	by
a diagram $d_\beta$, called a $(k,l)$--Brauer diagram,
	consisting of two rows of vertices and edges between vertices such that there are 
	1) $l$ vertices in the top row, labelled left to right by $1, \dots, l$;
	2) $k$ vertices in the bottom row, labelled left to right by $l+1, \dots, l+k$; and 
	3) the edges between vertices correspond to the blocks of $\beta$.
In particular, this means that each vertex is incident to exactly one edge; hence there are precisely $\frac{l+k}{2}$ edges in total in $d_{\beta}$. 

	\begin{comment}
For each $(k,l)$--Brauer partition $\beta$, we can represent it by
a diagram $d_\beta$, called a $(k,l)$--Brauer diagram,
which has a top row consisting of $l$ vertices and a bottom row consisting of $k$ vertices that has 
	$\frac{l+k}{2}$ edges between pairs of vertices, as follows.

Firstly, we need to label the vertices.
There are a number of different possible choices for how to label these vertices, but we make an intentional decision to label the top row, from left to right, by $1, \dots, l$ and the bottom row, again from left to right, by $l+1, \dots, l+k$.
We will see that such a labelling has been chosen to correspond to
	%matches nicely with 
	how the matrix elements of 
$\Hom_G((\mathbb{R}^{n})^{\otimes k}, (\mathbb{R}^{n})^{\otimes l})$
	are labelled.
Then, we draw an edge between any two vertices that appear in the same block in $\beta$.
This means that each vertex is incident to exactly one edge, and so there are precisely $\frac{l+k}{2}$ 
	edges in total in $d_{\beta}$, as desired.
	\end{comment}
\end{defn}

%We call the diagrams corresponding to each $(k,k)$--Brauer partition a $(k,k)$--Brauer diagram.

It is clear that the number of $(k,l)$--Brauer diagrams, if $l+k$ is even,
is $(l+k-1)!! \coloneqq (l+k-1)(l+k-3)\cdots5\cdot3\cdot1$, and is $0$ otherwise.

For example, we see that
\begin{equation} \label{brauerpartex1}
	\beta := \{1, 5 \mid 2, 8 \mid 3, 4 \mid 6, 7 \mid  9, 10\}
\end{equation}
is a valid $(6,4)$--Brauer partition. Figure \ref{brauerelement1}a) shows the $(6,4)$--Brauer diagram $d_{\beta}$ corresponding to $\beta$.
\begin{figure}[t]
	\begin{center}
	\scalebox{0.4}{\begin{tikzpicture}
	\begin{pgfonlayer}{nodelayer}
		\node [style=none] (94) at (3, 1) {$1$};
		\node [style=none] (95) at (6, 1) {$2$};
		\node [style=none] (96) at (9, 1) {$3$};
		\node [style=none] (97) at (12, 1) {$4$};
		\node [style=none] (98) at (0, -5) {$5$};
		\node [style=none] (99) at (3, -5) {$6$};
		\node [style=none] (100) at (6, -5) {$7$};
		\node [style=none] (101) at (9, -5) {$8$};
		\node [style=none] (102) at (12, -5) {$9$};
		\node [style=none] (103) at (15, -5) {$10$};
		\node [style=none] (104) at (-0.5, 0.5) {};
		\node [style=none] (105) at (-0.5, -4.5) {};
		\node [style=none] (106) at (15.5, -4.5) {};
		\node [style=none] (107) at (15.5, 0.5) {};
		\node [style=node] (108) at (3, 0) {};
		\node [style=node] (109) at (6, 0) {};
		\node [style=node] (110) at (9, 0) {};
		\node [style=node] (111) at (12, 0) {};
		\node [style=node] (112) at (0, -4) {};
		\node [style=node] (113) at (3, -4) {};
		\node [style=node] (114) at (6, -4) {};
		\node [style=node] (115) at (9, -4) {};
		\node [style=node] (116) at (12, -4) {};
		\node [style=node] (117) at (15, -4) {};
		\node [style=none] (118) at (24, 1) {$1$};
		\node [style=none] (119) at (27, 1) {$2$};
		\node [style=none] (120) at (30, 1) {$3$};
		\node [style=none] (121) at (33, 1) {$4$};
		\node [style=none] (122) at (21, -5) {$5$};
		\node [style=none] (123) at (24, -5) {$6$};
		\node [style=none] (124) at (27, -5) {$7$};
		\node [style=none] (125) at (30, -5) {$8$};
		\node [style=none] (126) at (33, -5) {$9$};
		\node [style=none] (127) at (36, -5) {$10$};
		\node [style=node] (132) at (24, 0) {};
		\node [style=node] (133) at (27, 0) {};
		\node [style=node] (134) at (30, 0) {};
		\node [style=node] (135) at (33, 0) {};
		\node [style=node] (136) at (21, -4) {};
		\node [style=node] (137) at (24, -4) {};
		\node [style=node] (138) at (27, -4) {};
		\node [style=node] (139) at (30, -4) {};
		\node [style=node] (140) at (33, -4) {};
		\node [style=node] (141) at (36, -4) {};
		\node [style=none] (142) at (20.5, 0.5) {};
		\node [style=none] (143) at (20.5, -4.5) {};
		\node [style=none] (144) at (36.5, -4.5) {};
		\node [style=none] (145) at (36.5, 0.5) {};
		\node [style=none] (146) at (-3, -2) {\huge{a)}};
		\node [style=none] (147) at (18, -2) {\huge{b)}};
	\end{pgfonlayer}
	\begin{pgfonlayer}{edgelayer}
		\draw [style={dash_2}] (105.center)
			 to (106.center)
			 to (107.center)
			 to (104.center)
			 to cycle;
		\draw [style=thick] (112) to (108);
		\draw [style=thick] (113) to (114);
		\draw [style=thick] (115) to (109);
		\draw [style=thick] (110) to (111);
		\draw [style=thick] (116) to (117);
		\draw [style={dash_3}] (143.center)
			 to (144.center)
			 to (145.center)
			 to (142.center)
			 to cycle;
		\draw [style=thick] (137) to (133);
		\draw [style=thick] (139) to (140);
	\end{pgfonlayer}
\end{tikzpicture}}
		\caption{a) The diagram $d_{\beta}$ corresponding to the $(6,4)$--Brauer partition $\beta$ given in (\ref{brauerpartex1}).
		b) The diagram $d_{\alpha}$ corresponding to the $(4+6) \backslash 6$--partition $\alpha$ given in (\ref{groodpartex1}).
		}
  	\label{brauerelement1}
	\end{center}
\end{figure}

The second type of diagram is what we have decided to call an $(l+k)\backslash n$--diagram
(pronounced ``$l$ \textit{plus} $k$ \textit{without} $n$").
These diagrams were originally hinted at by Brauer~\yrcite{Brauer} in the case where $l = k$ and were looked at in greater detail by Grood~\yrcite{grood}, but again only in the case where $l = k$. 
In their paper, they called these diagrams $k\backslash m$--diagrams, since they only considered the situation where $l = k$ and $n = 2m$. 
We will see that the definition below is equivalent to their definition in this case. 
Our naming convention for the diagrams makes more sense since they are a generalisation of those considered by Brauer and Grood.
%, we have chosen to call these diagrams $l+k\backslash n$--diagrams
%since they are a generalisation of those them in the next section, where we will see that our naming convention makes more sense in the general case.

\begin{defn}
	For any $k, l$ and $n \in \mathbb{Z}_{\geq 0}$, 
	%and for any $n=2m$,
	an $(l+k)\backslash n$--partition is a partitioning of the set $[l+k]$ with some $n$ elements removed into a disjoint union of pairs.
	Once again, each pair is called a block.
	
	An $(l+k)\backslash n$--diagram is the representation of an $(l+k)\backslash n$--partition in its diagram form, constructed in 
	%exactly the same way as for the Brauer diagrams above. 
	a similar way to the 
	$(k,l)$--Brauer diagrams above, where there is still a top row consisting of $l$ vertices and a bottom row consisting of $k$ vertices, but now there are only 
	$\frac{l+k-n}{2}$ edges between pairs of vertices.
	In this case, the $n$ vertices removed from the set $[l+k]$ will not be incident to any edge, and an edge exists between any two vertices that are in the same pair.
	We call the $n$ vertices whose labels have been removed from $[l+k]$ \textit{free} vertices.
\end{defn}
It is clear that if $n > l+k$, then no 
%such 
$(l+k)\backslash n$--diagrams
%diagrams 
exist.
Also, no such diagrams exist
if $n$ is odd and $l+k$ is even, or if $n$ is even and $l+k$ is odd. 

Otherwise, that is, if $n \leq l+k$ and either 
$n$ is even and $l+k$ is even, or  
$n$ is odd and $l+k$ is odd, then
the number of $(l+k)\backslash n$--diagrams
is $\binom{l+k}{n}(l+k-n-1)!!$,
since there are $\binom{l+k}{n}$ ways to pick $n$ free vertices, and 
$(l+k-n-1)!!$ ways to pair up the remaining $l + k - n$ vertices.

For example, we see that, if $k = 6$ and $l = 4$, then
\begin{equation} \label{groodpartex1}
	\alpha := \{2, 6 \mid 8, 9 \}
\end{equation}
is a $(4+6) \backslash 6$--partition. Figure \ref{brauerelement1}b) shows the $(4+6) \backslash 6$--diagram $d_{\alpha}$ corresponding to $\alpha$.

\begin{remark}
	We choose throughout to focus on $(k,l)$--Brauer and $(l+k)\backslash n$--\textit{diagrams} over their equivalent set partition form.
	%in the following.
	This is because both the diagrams and the matrices that they correspond to have matching shapes.
	%In particular, we chose to label the vertices as we did so that the labels in the top row correspond to the indices of a tuple in 
	%%because the shape of both the diagrams and the matrices that they correspond to are matching.
	%%the diagrams corresponds to the shape of the matrix that they correspond to.
	%%Notably, given how we chose to label the vertices (which was an intentional decision),
	%%the labels for the top row correspond to the indices of a tuple in 
	%$[n]^l$, and the labels for the bottom row correspond to the indices of a tuple in $[n]^k$. 
	%There are a number of different possible choices for how to label these vertices, but we make an intentional decision to label the top row, from left to right, by $1, \dots, l$ and the bottom row, again from left to right, by $l+1, \dots, l+k$.
	%We will see that such a labelling has been chosen to correspond to
	%%matches nicely with 
	%how the matrix elements of 
	In fact, it will become clear that, using these diagrams, we can view the matrix multiplication of a spanning set element in
	$\Hom_G((\mathbb{R}^{n})^{\otimes k}, (\mathbb{R}^{n})^{\otimes l})$ with an input vector in $(\mathbb{R}^{n})^{\otimes k}$
	as a process represented by the corresponding diagram, where
	%Specifically, 
	the input vector is passed into the bottom row of the diagram, and an output vector is returned from the top row.
	%one can think of the diagram as a process representing how its corresponding spanning set element in
	%$\Hom_G((\mathbb{R}^{n})^{\otimes k}, (\mathbb{R}^{n})^{\otimes l})$
	%takes a vector as input into the bottom row and outputs another vector in the top row.
\end{remark}

From the two types of diagrams defined above, we form the following two vector spaces.

\begin{defn}
	We define the Brauer vector space, $B_k^l(n)$,
	%The first is called the Brauer space $B_k^l(n)$, 
	which exists for any integer $n \in \mathbb{Z}_{\geq 1}$ and for any $k, l \in \mathbb{Z}_{\geq 0}$,
	as follows.
Let $B_0^0(n) \coloneqq \mathbb{R}$. 
Otherwise, define $B_k^l(n)$ to be the $\mathbb{R}$-linear span of the set of all $(k,l)$--Brauer diagrams.
\end{defn}

%When $l = k$, $B_k^k(n)$ can be given the structure of an algebra over $\mathbb{R}$, but as we are only interested in the individual elements of the basis of diagrams, we do not provide the details in this paper. See \cite{Brauer} for more details.

\begin{defn}
%The second is what we have decided to call 
	We define the Brauer--Grood vector space, $D_k^l(n)$, which exists for any integer $n \in \mathbb{Z}_{\geq 1}$ and for any $k,l \in \mathbb{Z}_{\geq 0}$,
	as follows.
Let $D_0^0(n) \coloneqq \mathbb{R}$. 
	Otherwise, define $D_k^l(n)$ to be the $\mathbb{R}$-linear span of the set of all $(k,l)$--Brauer diagrams together with the set of all $(l+k)\backslash n$--diagrams.
\end{defn}
Clearly, if $n > l+k$, or if 
$n$ is odd and $l+k$ is even, or if $n$ is even and $l+k$ is odd, 
then $D_k^l(n) = B_k^l(n)$.

With these two vector spaces, we are now able to find a spanning set of matrices for 
$\Hom_G((\mathbb{R}^{n})^{\otimes k}, (\mathbb{R}^{n})^{\otimes l})$
in the standard basis (for $G = O(n), SO(n)$) or the symplectic basis (for $G = Sp(n), n = 2m$) of $\mathbb{R}^{n}$.

\begin{comment}
In order to explicitly state what these spanning sets are, we first note that, for any $k,l \in \mathbb{Z}_{\geq 1}$, 
as a result of picking the standard/symplectic basis for each of 
$(\mathbb{R}^{n})^{\otimes k}$ 
and
$(\mathbb{R}^{n})^{\otimes l}$ 
appearing in
$\Hom((\mathbb{R}^{n})^{\otimes k}, (\mathbb{R}^{n})^{\otimes l})$,
the vector space
$\Hom((\mathbb{R}^{n})^{\otimes k}, (\mathbb{R}^{n})^{\otimes l})$
has a standard basis of matrix units
\begin{equation}
	\{E_{I,J}\}_{I \in [n]^l, J \in [n]^k}
\end{equation}
where $I$ is a tuple $(i_1, i_2, \dots, i_l) \in [n]^l$, 
$J$ is a tuple $(j_1, j_2, \dots, j_k) \in [n]^k$
and $E_{I,J}$ has a $1$ in the $(I,J)$ position and is $0$ elsewhere.

Note that if $k = 0$ and 
$l \in \mathbb{Z}_{\geq 1}$
, then the standard basis of matrix units for
$\Hom(\mathbb{R}, (\mathbb{R}^{n})^{\otimes l})$
is
\begin{equation}
	\{E_{I,1}\}_{I \in [n]^l}
\end{equation}
where $I$ is a tuple $(i_1, i_2, \dots, i_l) \in [n]^l$, 
and $E_{I,1}$ has a $1$ in the $(I,1)$ position and is $0$ elsewhere.

Similarly, if $l=0$ and
$k \in \mathbb{Z}_{\geq 1}$
, then the standard basis of matrix units for
$\Hom((\mathbb{R}^{n})^{\otimes k}, \mathbb{R})$
is
\begin{equation}
	\{E_{1,J}\}_{J \in [n]^k}
\end{equation}
where 
$J$ is a tuple $(j_1, j_2, \dots, j_k) \in [n]^k$
and $E_{1,J}$ has a $1$ in the $(1,J)$ position and is $0$ elsewhere.

Finally, if $k = l = 0$, then the standard basis of matrix units for
$\Hom(\mathbb{R}, \mathbb{R})$
is the $1 \times 1$ matrix
\begin{equation}
	\{E_{1,1} = (1) \}
\end{equation}
\end{comment}

In order to explicitly state what these spanning sets are, we note that, for any $k,l \in \mathbb{Z}_{\geq 0}$, 
as a result of picking the standard/symplectic basis for $\mathbb{R}^{n}$,
the vector space
$\Hom((\mathbb{R}^{n})^{\otimes k}, (\mathbb{R}^{n})^{\otimes l})$
has a standard basis of matrix units
\begin{equation} \label{standardbasisunits}
	\{E_{I,J}\}_{I \in [n]^l, J \in [n]^k}
\end{equation}
where $I$ is a tuple $(i_1, i_2, \dots, i_l) \in [n]^l$, 
$J$ is a tuple $(j_1, j_2, \dots, j_k) \in [n]^k$
and $E_{I,J}$ has a $1$ in the $(I,J)$ position and is $0$ elsewhere.
If one or both of $k$, $l$ is equal to $0$, then we replace the tuple that indexes the matrix by a $1$.
For example, when $k = 0$ and $l \in \mathbb{Z}_{\geq 1}$, (\ref{standardbasisunits}) becomes $\{E_{I,1}\}_{I \in [n]^l}$.

%Brauer's result for
%$\End_G((\mathbb{R}^{n})^{\otimes k})$
%where $G$ is either $O(n)$, $SO(n)$, or (for $n=2m$) $Sp(n)$,
%are summarised in the following three theorems.

We obtain the following results, which are given
in the following three theorems.

\begin{theorem} 
	%[Spanning set for 
	%$\Hom_{O(n)}((\mathbb{R}^{n})^{\otimes k}, (\mathbb{R}^{n})^{\otimes l})$] 
	[Spanning set when $G = O(n)$]
	\label{spanningsetO(n)}

	For any $k, l \in \mathbb{Z}_{\geq 0}$ and any
	$n \in \mathbb{Z}_{\geq 1}$, there is a surjection of vector spaces
	\begin{equation} \label{surjectionO(n)}
		\Phi_{k,n}^l : B_k^l(n) \rightarrow 
		\Hom_{O(n)}((\mathbb{R}^{n})^{\otimes k}, (\mathbb{R}^{n})^{\otimes l})
	\end{equation}
	which is defined as follows.

	If $l+k$ is odd, then we map the empty set onto the empty set. 
	Hence, in this case, 
		$\Hom_{O(n)}((\mathbb{R}^{n})^{\otimes k}, (\mathbb{R}^{n})^{\otimes l}) = \varnothing$.
	
	%Otherwise, assume that $l+k$ is even.

	Otherwise, for any $k,l \in \mathbb{Z}_{\geq 0}$ and for 
	each $(k,l)$--Brauer diagram $d_\beta$, associate the indices $i_1, i_2, \dots, i_l$ with the vertices in the top row of $d_\beta$, and $j_1, j_2, \dots, j_k$ with the vertices in the bottom row of $d_\beta$.
	Then, for any $n \in \mathbb{Z}_{\geq 1}$, define
	\begin{equation} \label{Ebeta}
		E_\beta 
		\coloneqq
		\sum_{I \in [n]^l, J \in [n]^k} 
		\delta_{r_1, u_1}
		\delta_{r_2, u_2}
		\dots
		\delta_{r_{\frac{l+k}{2}}, u_{\frac{l+k}{2}}}
		E_{I,J}
	\end{equation}
	where $r_1, u_1, \dots, r_{\frac{l+k}{2}}, u_{\frac{l+k}{2}}$ is any permutation of the indices $i_1, i_2, \dots, i_l, j_1, j_2, \dots, j_k$ such that the vertices corresponding to $r_p, u_p$ are in the same block of $\beta$.

	\begin{comment}
	If $k = 0$ and $l \in \mathbb{Z}_{\geq 2}$ is even, then for each $(0,l)$--Brauer diagram $\beta$, $E_\beta$ is defined to be
	\begin{equation}
		E_\beta 
		\coloneqq
		\sum_{I \in [n]^l}
		\delta_{r_1, u_1}
		\delta_{r_2, u_2}
		\dots
		\delta_{r_{\frac{l}{2}}, u_{\frac{l}{2}}}
		E_{I,1}
	\end{equation}
	where $r_1, u_1, \dots, r_{\frac{l}{2}}, u_{\frac{l}{2}}$ is any permutation of the indices $i_1, i_2, \dots, i_l$
	such that $r_p, u_p$ are in the same block of $\beta$.

	Similarly, if $l=0$ and $k \in \mathbb{Z}_{\geq 2}$ is even, then for each $(k,0)$--Brauer diagram $\beta$, $E_\beta$ is defined to be
	\begin{equation}
		E_\beta 
		\coloneqq
		\sum_{J \in [n]^k}
		\delta_{r_1, u_1}
		\delta_{r_2, u_2}
		\dots
		\delta_{r_{\frac{k}{2}}, u_{\frac{k}{2}}}
		E_{1,J}
	\end{equation}
	where $r_1, u_1, \dots, r_{\frac{k}{2}}, u_{\frac{k}{2}}$ is any permutation of the indices $j_1, j_2, \dots, i_k$
	such that $r_p, u_p$ are in the same block of $\beta$.

	Finally, if $k = l = 0$, then, for the single $(0,0)$--Brauer diagram $\beta$, $E_\beta$ is defined to be the $1 \times 1$ matrix $E_{1,1} = (1)$.
	\end{comment}

	The adapted version of Brauer's Invariant Argument, given in the Technical Appendix, shows that
	(\ref{Ebeta}) defines a bijective correspondence between
	the set of all $(k,l)$--Brauer diagrams and a spanning set for
	%in each case, $E_\beta$ is an element of 
	$\Hom_{O(n)}((\mathbb{R}^{n})^{\otimes k}, (\mathbb{R}^{n})^{\otimes l})$.

	Consequently, when $l+k$ is even, the surjection of vector spaces given in (\ref{surjectionO(n)})
	is defined by 
	\begin{equation}
		d_\beta \mapsto E_\beta
	\end{equation}
	for all $(k,l)$--Brauer diagrams $d_\beta$, and is extended linearly on the basis of such diagrams for $B_k^l(n)$.

	Hence the set
	\begin{equation} \label{klOnSpanningSet}
		\{E_\beta \mid d_\beta \text{ is a } (k,l) \text{--Brauer diagram} \}
	\end{equation}
	is a spanning set for
	$\Hom_{O(n)}((\mathbb{R}^{n})^{\otimes k}, (\mathbb{R}^{n})^{\otimes l})$
	in the standard basis of $\mathbb{R}^{n}$, 
	of size $0$ when $l+k$ is odd, and of size $(l+k-1)!!$ when $l+k$ is even.

	Lehrer and Zhang~\yrcite{LehrerZhang} showed that when $2n \geq l+k$, $\Phi_{k,n}^l$ is an isomorphism of vector spaces,
	and so the set (\ref{klOnSpanningSet})
	forms a basis of 
	$\Hom_{O(n)}((\mathbb{R}^{n})^{\otimes k}, (\mathbb{R}^{n})^{\otimes l})$
	in this case.
\end{theorem}

\begin{theorem} 
	%[Spanning set for 
	%$\Hom_{Sp(n)}((\mathbb{R}^{n})^{\otimes k}, (\mathbb{R}^{n})^{\otimes l})$, $n = 2m$] 
	[Spanning set when $G = Sp(n), n = 2m$]
	\label{spanningsetSp(n)}

	For any $k, l \in \mathbb{Z}_{\geq 0}$ and any
	$n \in \mathbb{Z}_{\geq 2}$ such that $n = 2m$, there is a surjection of vector spaces
	\begin{equation} \label{surjectionSp(n)}
		X_{k,n}^l : B_k^l(n) \rightarrow 
		\Hom_{Sp(n)}((\mathbb{R}^{n})^{\otimes k}, (\mathbb{R}^{n})^{\otimes l})
	\end{equation}
	which is defined as follows.

	If $l+k$ is odd, then we map the empty set onto the empty set. 
	Hence, in this case, 
		$\Hom_{Sp(n)}((\mathbb{R}^{n})^{\otimes k}, (\mathbb{R}^{n})^{\otimes l}) = \varnothing$.
	
	Otherwise, 
	%assume that $l+k$ is even.
	for any $k,l \in \mathbb{Z}_{\geq 0}$ and for 
	each $(k,l)$--Brauer diagram $d_\beta$, associate the indices $i_1, i_2, \dots, i_l$ with the vertices in the top row of $d_\beta$, and $j_1, j_2, \dots, j_k$ with the vertices in the bottom row of $d_\beta$.
	Then, for any $n \in \mathbb{Z}_{\geq 2}$, define
	\begin{equation} \label{Fbeta}
		F_\beta 
		\coloneqq
		\sum_{I, J} 
		\gamma_{r_1, u_1}
		\gamma_{r_2, u_2}
		\dots
		\gamma_{r_{\frac{l+k}{2}}, u_{\frac{l+k}{2}}}
		E_{I,J}
	\end{equation}
	where the indices $i_p, j_p$ range over $1, 1', \dots, m, m'$,
	where $r_1, u_1, \dots, r_{\frac{l+k}{2}}, u_{\frac{l+k}{2}}$ is any permutation of the indices $i_1, i_2, \dots, i_l, j_1, j_2, \dots, j_k$ such that the vertices corresponding to $r_p, u_p$ are in the same block of $\beta$, and
	%\begin{equation} \label{gammarpup}
	%	\gamma_{r_p, u_p} \coloneqq
	%	\begin{cases}
	%		\delta_{r_p, u_p} & \text{if the vertices corresponding to } r_p, u_p \text{ are in different rows of } d_\beta \\
	%		\epsilon_{r_p, u_p} & \text{if the vertices corresponding to } r_p, u_p \text{ are in the same row of } d_\beta
    	%	\end{cases}
	%\end{equation}
	\begin{equation} \label{gammarpup}
		\gamma_{r_p, u_p} \coloneqq
		\begin{cases}
			\delta_{r_p, u_p} & 
		\begin{array}{l}
			\text{if the vertices corresponding to } \\
			r_p, u_p \text{ are in different rows of } d_\beta
		\end{array} \\
		%	\text{if the vertices corresponding to } r_p, u_p \text{ are in different rows of } d_\beta \\
			\epsilon_{r_p, u_p} & 
		\begin{array}{l}
			\text{if the vertices corresponding to } \\
			r_p, u_p \text{ are in the same row of } d_\beta
		\end{array} \\
			%\text{if the vertices corresponding to } r_p, u_p \text{ are in the same row of } d_\beta
    		\end{cases}
	\end{equation}
	where $\epsilon_{r_p, u_p}$ was defined in (\ref{epsilondef1}) and (\ref{epsilondef2}).

		%\begin{array}{l}
			%\text{if the vertices corresponding to } \\
			%r_p, u_p \text{ are in different rows of } d_\beta
		%\end{array} \\

	\begin{comment}
	If $k = 0$ and $l \in \mathbb{Z}_{\geq 2}$ is even, then for each $(0,l)$--Brauer diagram $\beta$, $F_\beta$ is defined to be
	\begin{equation}
		F_\beta 
		\coloneqq
		\sum_{I} 
		\epsilon_{r_1, u_1}
		\epsilon_{r_2, u_2}
		\dots
		\epsilon_{r_{\frac{l}{2}}, u_{\frac{l}{2}}}
		E_{I,1}
	\end{equation}
	where the indices $i_p$ range over $1, 1', \dots, m, m'$, and
	$r_1, u_1, \dots, r_{\frac{l}{2}}, u_{\frac{l}{2}}$ is any permutation of the indices $i_1, i_2, \dots, i_l$
	such that $r_p, u_p$ are in the same block of $\beta$. 

	Similarly, if $l=0$ and $k \in \mathbb{Z}_{\geq 2}$ is even, then for each $(k,0)$--Brauer diagram $\beta$, $F_\beta$ is defined to be
	\begin{equation}
		F_\beta 
		\coloneqq
		\sum_{J} 
		\epsilon_{r_1, u_1}
		\epsilon_{r_2, u_2}
		\dots
		\epsilon_{r_{\frac{k}{2}}, u_{\frac{k}{2}}}
		E_{1,J}
	\end{equation}
	where the indices $j_p$ range over $1, 1', \dots, m, m'$, and
	$r_1, u_1, \dots, r_{\frac{k}{2}}, u_{\frac{k}{2}}$ is any permutation of the indices $j_1, j_2, \dots, i_k$
	such that $r_p, u_p$ are in the same block of $\beta$.

	Finally, if $k = l = 0$, then, for the single $(0,0)$--Brauer diagram $\beta$, $F_\beta$ is defined to be the $1 \times 1$ matrix $E_{1,1} = (1)$.
	\end{comment}

	The adapted version of Brauer's Invariant Argument, given in the Technical Appendix, shows that
	(\ref{Fbeta}) defines a bijective correspondence between
	the set of all $(k,l)$--Brauer diagrams and a spanning set for
	%, in each case, $F_\beta$ is an element of 
	$\Hom_{Sp(n)}((\mathbb{R}^{n})^{\otimes k}, (\mathbb{R}^{n})^{\otimes l})$. 
	%for $n = 2m$.

	Consequently, when $l+k$ is even, the surjection of vector spaces given in (\ref{surjectionSp(n)})
	is defined by 
	\begin{equation}
		d_\beta \mapsto F_\beta
	\end{equation}
	for all $(k,l)$--Brauer diagrams $d_\beta$, and is extended linearly on the basis of such diagrams for $B_k^l(n)$.

	Hence the set
	\begin{equation} \label{klSpnSpanningSet}
		\{F_\beta \mid d_\beta \text{ is a } (k,l) \text{--Brauer diagram} \}
	\end{equation}
	is a spanning set for
	$\Hom_{Sp(n)}((\mathbb{R}^{n})^{\otimes k}, (\mathbb{R}^{n})^{\otimes l})$, for $n = 2m$,
	in the symplectic basis of $\mathbb{R}^{n}$,
	of size $0$ when $l+k$ is odd, and of size $(l+k-1)!!$ when $l+k$ is even.

	Lehrer and Zhang~\yrcite{LehrerZhang} showed that when $n \geq l + k$, $X_{k,n}^l$ is an isomorphism of vector spaces,
	and so the set (\ref{klSpnSpanningSet})
	forms a basis of 
	$\Hom_{Sp(n)}((\mathbb{R}^{n})^{\otimes k}, (\mathbb{R}^{n})^{\otimes l})$, for $n = 2m$,
	in this case.
\end{theorem}

\begin{theorem} 
	%[Spanning set for 
	%$\Hom_{SO(n)}((\mathbb{R}^{n})^{\otimes k}, (\mathbb{R}^{n})^{\otimes l})$] 
	[Spanning set when $G = SO(n)$]
	\label{spanningsetSO(n)}

	For any $k, l \in \mathbb{Z}_{\geq 0}$ and any $n \in \mathbb{Z}_{\geq 1}$, we construct a surjection of vector spaces
	\begin{equation} \label{surjectionSO(n)}
		\Psi_{k,n}^l : D_k^l(n) \rightarrow 
		\Hom_{SO(n)}((\mathbb{R}^{n})^{\otimes k}, (\mathbb{R}^{n})^{\otimes l})
	\end{equation}
	as follows.

	If $n > l+k$, or if $n$ is odd and $l+k$ is even, or if $n$ is even and $l+k$ is odd, then we saw that $D_k^l(n) = B_k^l(n)$.
	Hence, in these cases, $\Psi_{k,n}^l = \Phi_{k,n}^l$, and so
	$
	\Hom_{SO(n)}((\mathbb{R}^{n})^{\otimes k}, (\mathbb{R}^{n})^{\otimes l})
	=
	\Hom_{O(n)}((\mathbb{R}^{n})^{\otimes k}, (\mathbb{R}^{n})^{\otimes l})$.
	\begin{comment}
	\textbf{POTENTIALLY REMOVE IF I KEEP SUMMARY LATER}
	Specifically, if $n > l+k$ and $l+k$ is even, or if $n \leq l+k$ and $n$ is odd and $l+k$ is even, then 
	$\Hom_{SO(n)}((\mathbb{R}^{n})^{\otimes k}, (\mathbb{R}^{n})^{\otimes l})$ is spanned by the images of $B_k^l(n)$ under $\Phi_{k,n}^l$, and so the set
	\begin{equation} \label{klSOnSpanningSet}
		\{E_\beta \mid \beta \text{ is a } (k,l) \text{--Brauer diagram} \}
	\end{equation}
	is a spanning set for
	$\Hom_{SO(n)}((\mathbb{R}^{n})^{\otimes k}, (\mathbb{R}^{n})^{\otimes l})$
	in the standard basis of $\mathbb{R}^{n}$.

	If $n > l+k$ and $l+k$ is odd, or if $n \leq l+k$ and $n$ is even and $l+k$ is odd, then no $(k,l)$--Brauer diagrams exist, and so
	$\Hom_{SO(n)}((\mathbb{R}^{n})^{\otimes k}, (\mathbb{R}^{n})^{\otimes l}) = \varnothing$.
	\textbf{UP TO HERE}
	\end{comment}

	Otherwise, that is, if $n \leq l+k$, and either $n$ is even and $l+k$ is even, or $n$ is odd and $l+k$ is odd, there exist
	$(l+k)\backslash n$--diagrams.
	%$d_\alpha$. 

	For each such diagram $d_\alpha$, again associate 
the indices $i_1, i_2, \dots, i_l$ with the vertices in the top row of $d_\alpha$, and $j_1, j_2, \dots, j_k$ with the vertices in the bottom row of $d_\alpha$.
	Suppose that there are $s$ free vertices in the top row. Then there are $n-s$ free vertices in the bottom row.
	Relabel the $s$ free indices in the top row (from left to right) by 
	$t_1, \dots, t_s$, and the $n-s$ free indices in the bottom row (from left to right) by $b_1, \dots, b_{n-s}$.

	Then, define
		$
		\chi
			\left(\begin{smallmatrix} 
				1 & 2 & \cdots & s & s+1 & \cdots & n\\
				t_1 & t_2 & \cdots & t_s & b_1 & \cdots & b_{n-s}
			\end{smallmatrix}\right)
		$
		as follows: it is 
		$0$ if the elements $t_1, \dots, t_s, b_1, \dots, b_{n-s}$ are not distinct, otherwise, it is
		$	
		\sgn
			\left(\begin{smallmatrix} 
				1 & 2 & \cdots & s & s+1 & \cdots & n\\
				t_1 & t_2 & \cdots & t_s & b_1 & \cdots & b_{n-s}
			\end{smallmatrix}\right)
		$,
		considered as a permutation of $[n]$.

	As a result, for any $n \in \mathbb{Z}_{\geq 1}$, define $H_\alpha$ to be
	\begin{equation} \label{SO(n)Halpha}
		\sum_{I \in [n]^l, J \in [n]^k} 
		\chi
			\left(\begin{smallmatrix} 
				1 & 2 & \cdots & s & s+1 & \cdots & n\\
				t_1 & t_2 & \cdots & t_s & b_1 & \cdots & b_{n-s}
			\end{smallmatrix}\right)
		\delta(r,u)
		E_{I,J}
	\end{equation}
	where
	\begin{equation}
	\delta(r,u) \coloneqq 
		\delta_{r_1, u_1}
		\delta_{r_2, u_2}
		\dots
		\delta_{r_{\frac{l+k-n}{2}}, u_{\frac{l+k-n}{2}}}
	\end{equation}
	Here, $r_1, u_1, \dots, r_{\frac{l+k-n}{2}}, u_{\frac{l+k-n}{2}}$
	is any permutation of the indices 
	\begin{equation}	
	\{i_1, \dots, i_l, j_1, \dots, j_k\} \backslash \{t_1, \dots, t_s, b_1, \dots, b_{n-s}\}
	\end{equation}
	such that the vertices corresponding to $r_p, u_p$ are in the same block of $\alpha$.

	The adapted version of Brauer's Invariant Argument, given in the Technical Appendix, shows that 
	(\ref{Ebeta}) and (\ref{SO(n)Halpha})
	defines a bijective correspondence between
	the set of all $(k,l)$--Brauer diagrams together with the set of all $(l+k) \backslash n$--diagrams,
	and a spanning set for
	%$H_\alpha$ is an element of 
	$\Hom_{SO(n)}((\mathbb{R}^{n})^{\otimes k}, (\mathbb{R}^{n})^{\otimes l})$.
	%when $n \leq l+k$, and either $n$ is even and $l+k$ is even, or $n$ is odd and $l+k$ is odd.

	Consequently, 
	when $n \leq l+k$ and 
	$n$ is even and $l+k$ is even, 
	the surjection of vector spaces (\ref{surjectionSO(n)})
	is given by
	\begin{equation}
		d_\beta \mapsto E_\beta
	\end{equation}
	if $d_\beta$ is a $(k,l)$--Brauer diagram, where $E_\beta$ was defined in Theorem \ref{spanningsetO(n)},
	and by
	\begin{equation} \label{surjdalpha}
		d_\alpha \mapsto H_\alpha
	\end{equation}
	if $d_\alpha$ is an $(l+k)\backslash n$--diagram, and is extended linearly on the basis of such diagrams for $D_k^l(n)$.
	
	When $n \leq l+k$ and $n$ is odd and $l+k$ is odd,
	the surjection of vector spaces (\ref{surjectionSO(n)})
	is given solely by (\ref{surjdalpha}), since no $(k,l)$--Brauer diagrams exist in this case.

	Hence, in all cases,
	%when $n \leq l+k$, and either $n$ is even and $l+k$ is even, or $n$ is odd and $l+k$ is odd,
	the set
	%\begin{equation} \label{SOn2SpanningSet}
		%\{E_\beta \mid \beta \text{ is a } (k,l) \text{--Brauer diagram} \} 
		%\cup
		%\{H_\alpha \mid \alpha \text{ is a } (k+l) \backslash n \text{--diagram} \} 
	%\end{equation}
	\begin{equation} \label{SOn2SpanningSet}
		\{E_\beta\}_{\beta}
		%\mid \beta \text{ is a } (k,l) \text{--Brauer diagram} \} 
		\cup
		\{H_\alpha\}_{\alpha}
		%\mid \alpha \text{ is a } (k+l) \backslash n \text{--diagram} \} 
	\end{equation}
	where $d_\beta$ is a $(k,l)$--Brauer diagram and $d_\alpha$ is an $(l+k) \backslash n$--diagram,
	is a spanning set for
	$\Hom_{SO(n)}((\mathbb{R}^{n})^{\otimes k}, (\mathbb{R}^{n})^{\otimes l})$
	in the standard basis of $\mathbb{R}^{n}$.

	\begin{comment}
	\textbf{FIX THIS TO MAKE IT LOOK GOOD FOR THE ONE COLUMN!}
	More explicitly, a spanning set for
	$\Hom_{SO(n)}((\mathbb{R}^{n})^{\otimes k}, (\mathbb{R}^{n})^{\otimes l})$
	in the standard basis of $\mathbb{R}^{n}$ is given by
	\begin{itemize}
		\item $n > l+k$
			\begin{itemize}
				\item $l+k$ odd: 
					the empty set $\varnothing$ of size $0$.	
				\item $l+k$ even: 
					the set
					$\{E_\beta \mid \beta \text{ is a } (k,l) \text{--Brauer diagram} \}$ of size $(k+l-1)!!$.
			\end{itemize}
		\item $n \leq l+k$
			\begin{itemize}
				\item $n$ odd, $l+k$ even: 
					the set
					$\{E_\beta \mid \beta \text{ is a } (k,l) \text{--Brauer diagram} \}$ of size $(k+l-1)!!$.
				\item $n$ even, $l+k$ odd: 
					the empty set $\varnothing$ of size $0$.	
				\item $n$ odd, $l+k$ odd: 
					the set
					$\{H_\alpha \mid \alpha \text{ is a } (k+l) \backslash n \text{--diagram} \}$ of size $\binom{l+k}{n}(l+k-n-1)!!$
				\item $n$ even, $l+k$ even: 
					the set
					$\{E_\beta \mid \beta \text{ is a } (k,l) \text{--Brauer diagram} \} 
		\cup
		\{H_\alpha \mid \alpha \text{ is a } (k+l) \backslash n \text{--diagram} \}$ of size $(k+l-1)!! + \binom{l+k}{n}(l+k-n-1)!!$
			\end{itemize}
	\end{itemize}
	\end{comment}
\end{theorem}

%We give examples for the results of Theorems 
%\ref{spanningsetO(n)}, \ref{spanningsetSp(n)}, and \ref{spanningsetSO(n)}
%in Appendix \ref{examples}.

\section{Adding Features and Biases} \label{featuresbiases}

\subsection{Features} 
In Section~\ref{SpanningSetResults},
we made the assumption that the feature dimension for all of the layers appearing in the neural network was one.
This simplified the analysis for the results seen in that section.
%However, 
All of these results can be adapted for the case where the feature dimension of the layers is greater than $1$.

%It is enough to note that if a $k$-order tensor has a feature space of dimension $d_k$, then the space under consideration is
%$(\mathbb{R}^{n})^{\otimes k} \otimes \mathbb{R}^{d_k}$.

%Consequently, we can adapt 
%Theorems \ref{spanningsetO(n)}, \ref{spanningsetSp(n)}, and \ref{spanningsetSO(n)} 
%of Section~\ref{SpanningSetResults}
%as follows. 

Suppose that an $r$-order tensor has a feature space of dimension $d_r$.
%, and an $l$-order tensor has a feature space of dimension $d_l$.
We now wish to find a spanning set for 
\begin{equation} \label{Homfeatures}
	\Hom_{G}((\mathbb{R}^{n})^{\otimes k} \otimes \mathbb{R}^{d_k}, (\mathbb{R}^{n})^{\otimes l} \otimes \mathbb{R}^{d_l})
\end{equation}
in the standard basis of $\mathbb{R}^{n}$ for $G = O(n)$ and $SO(n)$, and in the symplectic basis of $\mathbb{R}^{n}$ for $G = Sp(n)$, where $n = 2m$.
	
A spanning set in each case can be found by making the following substitutions
in Theorems \ref{spanningsetO(n)}, \ref{spanningsetSp(n)}, and \ref{spanningsetSO(n)} 
of Section~\ref{SpanningSetResults},
%in the results for Theorems \ref{spanningsetO(n)}, \ref{spanningsetSp(n)}, and \ref{spanningsetSO(n)}, 
where now $i \in [d_l]$ and $j \in [d_k]$:
\begin{itemize}
	\item replace $E_{I,J}$ by $E_{I,i,J,j}$, $E_{I,1}$ by $E_{I,i,1,j}$, $E_{1,J}$ by $E_{1,i,J,j}$, and $E_{1,1}$ by $E_{1,i,1,j}$, and
	\item relabel $E_\beta$ by $E_{\beta, i, j}$, $F_\beta$ by $F_{\beta, i, j}$, and $H_\alpha$ by $H_{\alpha, i, j}$.
\end{itemize}
Consequently, a spanning set for (\ref{Homfeatures})
	%$\Hom_{G}((\mathbb{R}^{n})^{\otimes k} \otimes \mathbb{R}^{d_k}, (\mathbb{R}^{n})^{\otimes l} \otimes \mathbb{R}^{d_l})$, 
	in the standard/symplectic basis of $\mathbb{R}^{n}$, is given by
%\begin{itemize}
	%\item 
		%$\{E_{\beta, i, j} \mid \beta \text{ is a } (k,l) \text{--Brauer diagram}, i \in [d_l], j \in [d_k] \}$, for $G = O(n)$, 
	%\item 
		%$\{F_{\beta, i, j} \mid \beta \text{ is a } (k,l) \text{--Brauer diagram}, i \in [d_l], j \in [d_k] \}$, for $G = Sp(n)$ (where $n = 2m$), and
	%\item 
		%$\{E_{\beta, i, j} \mid \beta \text{ is a } (k,l) \text{--Brauer diagram}, i \in [d_l], j \in [d_k] \} 
		%\cup
		%\{H_{\alpha, i, j} \mid \alpha \text{ is a } (k+l) \backslash n \text{--diagram}, i \in [d_l], j \in [d_k] \}$, for $G = SO(n)$.
%\end{itemize}
\begin{itemize}
	\item 
		$\{E_{\beta, i, j}\}_{\beta, i, j}$ for $G = O(n)$, 
	\item 
		$\{F_{\beta, i, j}\}_{\beta, i, j}$ for $G = Sp(n)$ (where $n = 2m$), and
	\item 
		$\{E_{\beta, i, j}\}_{\beta, i, j}
		\cup
		\{H_{\alpha, i, j}\}_{\alpha, i, j}$ for $G = SO(n)$,
\end{itemize}
where  
$d_\alpha$ is an $(l+k) \backslash n$--diagram,
$d_\beta$ is a $(k,l)$--Brauer diagram, $i \in [d_l]$, and $j \in [d_k]$.

\subsection{Biases} 
Including bias terms in the layer functions of a $G$-equivariant neural network is harder, but it can be done.
For the learnable linear layers of the form
$\Hom_{G}((\mathbb{R}^{n})^{\otimes k}, (\mathbb{R}^{n})^{\otimes l})$,
Pearce--Crump~\yrcite{pearcecrump} shows that the $G$-equivariance of the bias function, 
$\beta : ((\mathbb{R}^{n})^{\otimes k}, \rho_k) \rightarrow 
((\mathbb{R}^{n})^{\otimes l}, \rho_l)$,
needs to satisfy
	\begin{equation} \label{Gbiasequiv}
		c = \rho_l(g)c
	\end{equation}
for all $g \in G$ and $c \in (\mathbb{R}^{n})^{\otimes l}$.

Since any $c \in (\mathbb{R}^{n})^{\otimes l}$
satisfying (\ref{Gbiasequiv}) can be viewed as an element of 
$\Hom_{G}(\mathbb{R}, (\mathbb{R}^{n})^{\otimes l})$,
to find the matrix form of $c$, all we need to do is to find a spanning set for 
$\Hom_{G}(\mathbb{R}, (\mathbb{R}^{n})^{\otimes l})$.

But this is simply a matter of applying the results of Section~\ref{SpanningSetResults}, namely
Theorem \ref{spanningsetO(n)} for $G = O(n)$,
Theorem \ref{spanningsetSp(n)} for $G = Sp(n)$, with $n = 2m$, and
Theorem \ref{spanningsetSO(n)} for $G = SO(n)$,
setting $k = 0$.

\section{Equivariance to Local Symmetries} \label{EquivLocal}

We can extend our results to looking at linear layer functions that are equivariant to a direct product of groups. 
In this scenario, the data is given on a collection of some $p$ sets of sizes $n_1, \dots, n_p$, and we require 
equivariance to the group $G(n_i)$ for the data given on the $i^{\text{th}}$ set.
%that the data given on the $i^{\text{th}}$ set is equivariant to the group $G(n_i)$.
%We say that 
Neural networks that are constructed using these layer functions are said to be equivariant to local symmetries.
%that is, we can construct neural networks that are equivariant to local symmetries.

Specifically, we wish to find a spanning set for 
%\begin{equation} \label{genericHomspace}
	%\Hom_{G(n_1) \times \dots \times G(n_p)}((\mathbb{R}^{n_1})^{\otimes {k_1}} \boxtimes \dots \boxtimes (\mathbb{R}^{n_p})^{\otimes {k_p}}, 
%(\mathbb{R}^{n_1})^{\otimes {l_1}} \boxtimes \dots \boxtimes (\mathbb{R}^{n_p})^{\otimes {l_p}})
%\end{equation}
\begin{equation} \label{genericHomspace}
	\Hom_{G(n_1) \times \dots \times G(n_p)}(V, W)
\end{equation}
where 
\begin{equation}
	V \coloneqq (\mathbb{R}^{n_1})^{\otimes {k_1}} \boxtimes \dots \boxtimes (\mathbb{R}^{n_p})^{\otimes {k_p}}
\end{equation}
\begin{equation}
	W \coloneqq (\mathbb{R}^{n_1})^{\otimes {l_1}} \boxtimes \dots \boxtimes (\mathbb{R}^{n_p})^{\otimes {l_p}}
\end{equation}
and $\boxtimes$ is the external tensor product.

The $\Hom$-space given in (\ref{genericHomspace}) is isomorphic to
\begin{equation} \label{tensorgenericHomspace}
	\bigotimes_{r=1}^{p} \Hom_{G(n_r)}((\mathbb{R}^{n_r})^{\otimes k_r}, (\mathbb{R}^{n_r})^{\otimes l_r})
\end{equation}
Consequently, we can construct a surjection of vector spaces, denoted by
$\bigotimes_{r=1}^{p} \Theta_{{k_r},{n_r}}^{l_r}$, from 
$\bigotimes_{r=1}^{p} A_{k_r}^{l_r}(n_r)$ to the $\Hom$--space given in 
(\ref{tensorgenericHomspace}),
%\begin{equation}
	%\bigotimes_{r=1}^{p} \Theta_{{k_r},{n_r}}^{l_r} :
	%\bigotimes_{r=1}^{p} A_{k_r}^{l_r}(n_r)
	%\rightarrow
	%\bigotimes_{r=1}^{p} \Hom_{G(n_r)}((\mathbb{R}^{n_r})^{\otimes k_r}, (\mathbb{R}^{n_r})^{\otimes l_r})
%\end{equation}
where 
\begin{itemize}
	\item if $G(n_r) = O(n_r)$, then $\Theta_{{k_r},{n_r}}^{l_r} = \Phi_{{k_r},{n_r}}^{l_r}$ and $A_{k_r}^{l_r}(n_r) = B_{k_r}^{l_r}(n_r)$,
	\item if $G(n_r) = Sp(n_r)$, then $\Theta_{{k_r},{n_r}}^{l_r} = X_{{k_r},{n_r}}^{l_r}$ and $A_{k_r}^{l_r}(n_r) = B_{k_r}^{l_r}(n_r)$ (here $n_r = 2m_r$), and
	\item if $G(n_r) = SO(n_r)$, then $\Theta_{{k_r},{n_r}}^{l_r} = \Psi_{{k_r},{n_r}}^{l_r}$ and $A_{k_r}^{l_r}(n_r) = D_{k_r}^{l_r}(n_r)$.
\end{itemize}
As a result, a spanning set for (\ref{genericHomspace})
can be found by placing each possible basis diagram for each of the vector spaces $A_{k_r}^{l_r}(n_r)$
side by side, taking the image of each under its map
$\Theta_{{k_r},{n_r}}^{l_r}$, and then calculating the Kronecker product of the resulting matrices.
%to obtain the spanning set for (\ref{genericHomspace}).

Features and biases can be added in exactly the same way as discussed in Section~\ref{featuresbiases}.

\section{Related Work}

The combinatorial representation theory of the Brauer algebra was developed by Brauer~\yrcite{Brauer}
for the purpose of understanding the centraliser algebras of the groups $O(n)$, $SO(n)$ and $Sp(n)$.
Brown published two papers~\yrcite{brown1,brown2}
%in the 1950s 
on the Brauer algebra,
%, written as $B_k^k(n)$ in our notation, 
showing that it is semisimple if and only if $n \geq k-1$.
Weyl~\yrcite{weyl} had previously shown that the Brauer algebra was semisimple if $n \geq 2k$.
After Brown's papers, the Brauer algebra was largely forgotten about until Hanlon and Wales~\yrcite{hanlonwales} provided an isomorphism between two versions of the Brauer algebra -- these versions share a common basis but have different algebra products defined on them.
Grood~\yrcite{grood} investigated the representation theory of what we have termed the Brauer--Grood algebra.
%, written as $D_k^k(n)$ in our notation, 
%as \cite{Brauer} only contained half a page on the algebra and nothing more had been published on it since its publication.
Lehrer and Zhang~\yrcite{LehrerZhang} studied the kernel of the surjection of algebras given in Theorems \ref{spanningsetO(n)} and \ref{spanningsetSp(n)} 
%of Section~\ref{SpanningSetResults}
when $l = k$, showing that, in each case, it is a two-sided ideal generated by a single element of the Brauer algebra.

With regard to the machine learning literature, Maron et al.'s paper~\yrcite{maron2018} is the closest to ours in terms of how it motivated our research idea.
They characterised all of the learnable, linear, equivariant layer functions when the layers are some tensor power of $\mathbb{R}^{n}$ for the symmetric group $S_n$ in the practical cases (specifically, when $n \geq k+l$). 
Pearce--Crump~\yrcite{pearcecrump} used the Schur--Weyl duality between the symmetric group and the partition algebra to provide a full characterisation for these layer functions for all values of $n$ and for all orders of the tensor power spaces involved, showing, in particular, that the dimension of the space of layer functions is not independent of $n$.
%However, as we show in our paper \cite{pearcecrump}, their characterisation is not entirely correct, since they conclude, incorrectly, that the dimension of the layer functions is independent of $n$.
Finzi et al.~\yrcite{finzi} were the first to recognise that the dimension is not independent of $n$,
using a numerical algorithm to calculate the correct values for small values of $n$, $k$ and $l$.
%noting this in the appendix of their paper.
%In 2021, Finzi et al. \cite{finzi} were the first to recognise Maron et al.'s error, noting this in the appendix of their paper. 
%However, they were only able to calculate the correct values for the dimension for a few values of $k$ and $n$ as their approach uses a numerical algorithm which runs out of memory on higher values of $k$ and $n$.
%In our paper, we provide an analytic solution for these dimensions, for any $k$ and $n$, fully correcting Maron et al..
Their numerical algorithm also enabled them to find a basis for the learnable, layer, equivariant functions for the groups that are the focus of our study, namely $O(n)$, $Sp(n)$ and $SO(n)$, but only for small values of $n, k$ and $l$, since their algorithm 
%once again 
runs out of memory on higher values.
In this paper, whilst we have not found a basis in all cases, we have provided a spanning set and an analytic solution for all values of $n, k$ and $l$, which will make it possible to implement
%enable the implementation of
%be good enough to implement 
group equivariant neural networks for any such values of $n, k$ and $l$ for the three groups in question.
In writing up this paper, we came across a paper by Villar et al.~\yrcite{villar2021scalars}, in which they focus on designing group equivariant neural networks for $O(n)$ and $SO(n)$, amongst others.
They also use the First Fundamental Theorem of Invariant Theory for $O(n)$ and $SO(n)$, but only to characterise all invariant scalar functions 
$(\mathbb{R}^{n})^{\otimes k} \rightarrow \mathbb{R}$ 
and all equivariant vector functions
$(\mathbb{R}^{n})^{\otimes k} \rightarrow \mathbb{R}^{n}$
as a sum involving $O(n)$ or $SO(n)$ invariant scalar functions.
They then use multilayer perceptrons to learn these invariant scalar functions.
Our method, by contrast, characterises a wider selection of functions, since we study the linear, learnable, equivariant functions between 
layers that are any tensor power of $\mathbb{R}^{n}$ for $O(n)$ and $SO(n)$ (as well as for $Sp(n)$), and
we give the exact matrix form of these functions in the standard basis of $\mathbb{R}^{n}$, meaning that the group equivariant neural network architectures that we provide are exact and do not require any approximations via multilayer perceptrons.

\section{Limitations, Discussion, and Future Work}

We believe that our approach for characterising all of the possible
$O(n), Sp(n)$ and $SO(n)$ equivariant neural networks whose layers are some tensor power of $\mathbb{R}^{n}$
is promising in terms of the possible practical applications.
A number of papers have appeared recently in the literature where the authors tried to learn group equivariant functions -- for the groups given in this paper -- on tensor power spaces of $\mathbb{R}^{n}$. 
However, they were not especially successful in their attempts.
%outcomes contained within them have not been especially successful. 
We believe that this is a consequence of them using architectures 
that approximate 
the group equivariance property of the functions that they wish to learn,
rather than guaranteeing it exactly.
%being approximate rather than exact in terms of guaranteeing 
%%them having approximate rather than exact architectures to guarantee 
%the group equivariance of the neural networks.
The results in this paper directly address this problem.
In Finkelshtein et al.~\yrcite{finkelshtein},
the authors 
%looked to create 
created a tensor product neural network 
which was approximately equivariant to $O(3) \times S_n$ in order
to learn from point cloud data.
%which involved tensor power spaces for the $O(3)$ group. 
%As the authors themselves state in their paper, their attempt was not particularly successful. 
%However, w
By combining the results of Pearce--Crump~\yrcite{pearcecrump}
for $S_n$ with the results in this paper for $O(3)$, we would be able to replace the 
%ascending and descending 
linear layers in their architecture
%Finkelshtein et al.~\yrcite{finkelshtein}
with exact, parameterizable matrices for the tensor product spaces that are guaranteed to be equivariant to $O(3) \times S_n$.
We believe that this could improve the final outcome. 
In addition, in Villar et al.~\yrcite{villar2021scalars},
the authors explore two numerical experiments involving tensor power representations: the first is an $O(5)$-invariant task from an order $2$ tensor power of $\mathbb{R}^5$ to $\mathbb{R}$, and the second is an $S_5 \times O(3)$ task where the $O(3)$ component is equivariant on order $5$ tensors of $\mathbb{R}^3$. 
The authors, however, use standard multi-layer perceptrons to learn the functions. 
We think that they could improve upon their results by using
%replacing the multi-layer perceptrons with 
the linear layers that are characterised in this paper. 
%As an additional point
Furthermore, we are of the opinion that by characterising the equivariant neural networks for these groups, we have made it possible for other researchers in the machine learning community to find further uses for these neural networks.

We are aware that equivariance to the symplectic group $Sp(n)$ does not commonly appear in the machine learning literature.
However, as stated in Appendix E.6 of Finzi et al.~\yrcite{finzi},
$Sp(n)$ equivariance is especially relevant in the context of Hamiltonian mechanics and classical physics. 
Section 7.2 of Finzi et al.~\yrcite{finzi} points to the paper by Greydanus et al.~\yrcite{greydanus},
where the authors look to learn the Hamiltonian of a system coming from Hamiltonian mechanics.
In particular, the time evolution of the system is expressed in terms of the symplectic basis. 
The paper by Villar et al.~\yrcite{villar2021scalars} also highlights how there are many symmetries in physics that are relevant for the machine learning community, including “symplectic symmetry that permits reinterpretations of positions and momenta”.

We appreciate that given the current state of hardware, there will be some computational limitations when implementing the neural networks that appear in this paper in practice, and some engineering may be required to obtain the necessary scale.
In particular, it is not a trivial task to store the high order tensors that appear in such neural networks.
This was demonstrated by Kondor et al.~\yrcite{clebschgordan},
where the authors needed to develop custom CUDA kernels in order to implement their tensor product based neural networks.
In saying that, however, we feel that given the ever-increasing availability of computing power, we should see higher-order group equivariant neural networks appear more often in practice. 
We also note that while the tensor power spaces increase exponentially in dimension as we increase their order, the dimension of the space of equivariant maps between such tensor spaces is typically much smaller, and the matrices themselves are often sparse.
Hence, while there may be some technical difficulties in storing such matrices, it should be possible to do so with the current computing power that is available.

We recognise, however, that further work is required in order to demonstrate fully the practical applicability of our results. In particular, we need to conduct practical experiments to assess how these neural networks perform when the order of the tensor power spaces is increased, and we need to show that these neural networks provide sufficient practical advantages over the existing approaches that use irreducible decompositions.
%the existing approaches.

\section{Conclusion} \label{conclusion}

We are the first to show how the combinatorics underlying
the Brauer and Brauer--Grood vector spaces provides the theoretical background for constructing group equivariant neural networks for the orthogonal, special orthogonal, and symplectic groups
when the layers are some tensor power of $\mathbb{R}^{n}$.
We looked at the problem of calculating the matrix form of the linear layer functions between such spaces in the standard/symplectic basis for $\mathbb{R}^{n}$.
We recognised that a solution to this problem would provide a powerful method for constructing
group equivariant neural networks for the three groups in question since we could avoid having to solve the much more difficult problem of decomposing the 
tensor power spaces of $\mathbb{R}^{n}$ into their irreducible representations
%Consequently, we could 
and then avoid having to find the change of basis matrix that would be needed
to perform the layer mappings.

We saw how a basis of diagrams for the Brauer and Brauer--Grood vector spaces enabled us to find a spanning set 
of matrices for the layer functions themselves in the standard/symplectic basis for $\mathbb{R}^{n}$
%for
%$\Hom_{G}((\mathbb{R}^{n})^{\otimes k}, (\mathbb{R}^{n})^{\otimes l})$ 
for each of the three groups in question, and how each diagram provided the parameter sharing scheme for its image in the spanning set.
As a result, we were able to characterise all of the possible group equivariant neural networks 
whose layers are some tensor power of $\mathbb{R}^{n}$
for each of the three groups in question.
We were also able to generalise this diagrammatic approach to layer functions that were equivariant to local symmetries.

As discussed in the Introduction, our results were motivated by
Brauer~\yrcite{Brauer} who showed that
%, for each of the three groups in question, 
there exists a Schur--Weyl duality for each of the three groups in question
%for each of the three groups in question with 
with an algebra of diagrams.
\begin{comment}
\textbf{REDUCE THIS DOWN TO FIT}
As discussed in the Introduction, one of the major motivating factors behind how we came to these results is that
%if we set $l = k$, then 
%it can be shown that
Brauer showed that 
$O(n)$ is in Schur--Weyl duality with the Brauer algebra $B_k^k(n)$,
$Sp(n)$ is in Schur--Weyl duality with the Brauer algebra $B_k^k(n)$,
and
$SO(n)$ is in Schur--Weyl duality with the Brauer--Grood algebra $D_k^k(n)$, since
the surjection of vector spaces that we have constructed in
Theorems \ref{spanningsetO(n)}, \ref{spanningsetSp(n)}, and \ref{spanningsetSO(n)}
%, namely
%$B_k^k(n) \rightarrow \End_{O(n)}((\mathbb{R}^{n})^{\otimes k})$,
%$B_k^k(n) \rightarrow \End_{Sp(n)}((\mathbb{R}^{n})^{\otimes k})$ ($n = 2m$), and
%$D_k^k(n) \rightarrow \End_{SO(n)}((\mathbb{R}^{n})^{\otimes k})$,
actually become a surjection of algebras when $l = k$.
\end{comment}
Consequently, this leads to the following question, which is one for future research: what other Schur--Weyl dualities exist between a group and an algebra of diagrams that would enable us to understand the structure of neural networks that are equivariant to the group?
%For example, in a separate paper
%As discussed above, we have shown \cite{pearcecrump} that the Schur--Weyl duality that exists between the symmetric group, $S_n$, and 
%the partition algebra, $P_k(n)$, enables us to fully characterise all of the possible permutation equivariant neural networks when the layers are some tensor power of $\mathbb{R}^{n}$.

\section*{Acknowledgements}

The author would like to thank his PhD supervisor Professor William J. Knottenbelt for being generous with his time throughout the author's period of research prior to the publication of this paper.

This work was funded by the Doctoral Scholarship for Applied Research which was awarded to the author under Imperial College London's Department of Computing Applied Research scheme.
This work will form part of the author's PhD thesis at Imperial College London.

%%%%%%%%%%%%%%%%%%%%%%%%%%%%%%%%%%%%%%%%%%%%%%%%%%%%%%%%%%%%%%%%%%%%%%%%%%%%%%%%%%%%%%%%%%

\newpage

\nocite{*}
\bibliography{references}
\bibliographystyle{icml2023}

%%%%%%%%%%%%%%%%%%%%%%%%%%%%%%%%%%%%%%%%%%%%%%%%%%%%%%%%%%%%%%%%%%%%%%%%%%%%%%%
%%%%%%%%%%%%%%%%%%%%%%%%%%%%%%%%%%%%%%%%%%%%%%%%%%%%%%%%%%%%%%%%%%%%%%%%%%%%%%%
% APPENDIX
%%%%%%%%%%%%%%%%%%%%%%%%%%%%%%%%%%%%%%%%%%%%%%%%%%%%%%%%%%%%%%%%%%%%%%%%%%%%%%%
%%%%%%%%%%%%%%%%%%%%%%%%%%%%%%%%%%%%%%%%%%%%%%%%%%%%%%%%%%%%%%%%%%%%%%%%%%%%%%%
\newpage
\appendix
\onecolumn

\section{Brauer's Invariant Argument, adapted for 
	$\Hom_G((\mathbb{R}^{n})^{\otimes k}, (\mathbb{R}^{n})^{\otimes l})$
	} \label{BrauerInvariantArgument}

\subsection{Some Preliminary Material}
%ahead of Brauer's Invariant Argument}

We consider throughout the real vector space $\mathbb{R}^{n}$.
%of dimension $n$, for $n \in \mathbb{Z}_{\geq 1}$.

Let $GL(n)$ be the group of invertible linear transformations from 
$\mathbb{R}^{n}$ to $\mathbb{R}^{n}$.
Let $G$ be any subgroup of $GL(n)$.

Recall that the vector space 
$\mathbb{R}^{n}$ 
	has, associated with it, its dual vector space, 
	$(\mathbb{R}^{n})^{*}$.
Let $B \coloneqq \{b_i \mid i \in [n]\}$ be any basis of 
$\mathbb{R}^{n}$.
It has associated with it the dual basis $B^{*} \coloneqq \{b_i^{*} \mid i \in [n]\}$, a basis of 
	$(\mathbb{R}^{n})^{*}$,
	such that $b_i^{*}(b_j) = \delta_{i,j}$.

In particular, coordinates on 
$\mathbb{R}^{n}$
with respect to the basis $B$ are linear functions, that is, elements of 
	$(\mathbb{R}^{n})^{*}$.
Indeed, if $v = \sum_{j} x_jb_j$, then the coordinate function $x_j$ can be identified with the dual basis vector $b_j^{*}$ since
\begin{equation}
	b_j^{*}(v) 
	= b_j^{*}(\sum_{i} x_ib_i) 
	= \sum_{i} x_ib_j^{*}(b_i) 
	= x_j
\end{equation}
Since $G$ is a subgroup of $GL(n)$, we see that 
$\mathbb{R}^{n}$
is a representation of $G$, which we denote by $\rho_1$ in the following.
In fact, for all $f \in G$, we have that $\rho_1(f) = f$.

Consequently, if $v$ is any vector in 
$\mathbb{R}^{n}$,
and $f$ is any element of $G$, then the linear transformation
\begin{equation}
	\rho_1(f) = f : v \rightarrow f(v)
\end{equation}
can be expressed in its matrix representation, choosing $B$ as the basis for each copy of 
$\mathbb{R}^{n}$,
as the matrix $a(f) = (a_{i,j})$ such that
\begin{equation} \label{rho1action}
	y_i = \sum_{j} a_{i,j} x_j		
\end{equation}
where, in the basis $B$,
\begin{equation} \label{rho1basisvecs}
	v = \sum_{j} x_jb_j \text{ and } f(v) = \sum_{j} y_jb_j
\end{equation}
for some coefficients $x_j, y_j \in \mathbb{R}$.

We will sometimes express (\ref{rho1action}) in the form $y = a(f)x$, where $x, y$ are column vectors such that the $i^\text{th}$ component of each vector is $x_i$ and $y_i$, respectively.

As $(\mathbb{R}^{n}, \rho_1)$ is a representation of $G$, we can define another representation of $G$, called the contragredient representation, on the dual space 
$(\mathbb{R}^{n})^{*}$,
as follows
\begin{equation}
	(\rho_1^{-1})^\top : G \rightarrow GL((\mathbb{R}^{n})^{*})
\end{equation}
where, for all $f \in G$,
\begin{equation}
	(\rho_1^{-1})^\top(f) : (\mathbb{R}^{n})^{*} \rightarrow (\mathbb{R}^{n})^{*}
\end{equation}
is defined by
\begin{equation}
	(\rho_1^{-1})^\top(f)[u] : v \mapsto u(\rho_1^{-1}(f)(v))
\end{equation}

One way of understanding the contragredient representation 
$((\mathbb{R}^{n})^{*}, (\rho_1^{-1})^\top)$ 
of $G$ is as the action on 
$(\mathbb{R}^{n})^{*}$
such that all pairings between 
$(\mathbb{R}^{n})^{*}$
and
$\mathbb{R}^{n}$
remain invariant under their respective actions.
Indeed, if $u \in (\mathbb{R}^{n})^{*}$,
and $v \in \mathbb{R}^{n}$,
then we see that, for all $f \in G$
\begin{equation}
	v \mapsto \rho_1(f)(v)
\end{equation}
and
\begin{equation}
	u \mapsto (\rho_1^{-1})^\top(f)[u]
\end{equation}
and so
\begin{equation} \label{contrainvariant}
	u(v) \mapsto (\rho_1^{-1})^\top(f)[u](\rho_1(f)(v)) = u(\rho_1^{-1}(f)(\rho_1(f)(v))) = u(v)
\end{equation}

Hence, expressing $u \in (\mathbb{R}^{n})^{*}$
in the dual basis $B^{*}$ as
\begin{equation} \label{contragredbasisvec}
	u = \sum_{j} p_jb_j^{*} 
\end{equation}
and $(\rho_1^{-1})^\top(f)[u] \in (\mathbb{R}^{n})^{*}$
as
\begin{equation}
	(\rho_1^{-1})^\top(f)[u] = \sum_{j} q_jb_j^{*}
\end{equation}
we see that, for any $f \in G$, the linear transformation
\begin{equation}
	(\rho_1^{-1})^\top(f) : u \rightarrow (\rho_1^{-1})^\top(f)[u]
\end{equation}
can be expressed in its matrix representation as
\begin{equation}
	pa(f)^{-1} = q
\end{equation}
as a result of (\ref{contrainvariant}),
and so
\begin{equation} \label{contragredientvecs}
	p = qa(f)
\end{equation}
that is,
\begin{equation} \label{contragredsum}
	p_i = \sum_{j} q_ja_{j,i}
\end{equation}
In (\ref{contragredientvecs}),
$p, q$ are row vectors such that the $i^\text{th}$ component of each is $p_i$ and $q_i$, respectively.

We also have that $(\mathbb{R}^{n})^{\otimes k}$
is a representation of $G$, which we denote by $\rho_k$. 
%where $\rho_k \coloneqq \rho_1^{\otimes k}$.
In particular, for all $f \in G$, we see that $\rho_k(f) = \rho_1(f)^{\otimes k} = f^{\otimes k}$.

Consequently, if $v$ is any vector in 
$(\mathbb{R}^{n})^{\otimes k}$,
and $f$ is any element of $G$, then the linear transformation
\begin{equation} \label{kpowerrep}
	\rho_k(f) = f^{\otimes k} : v \rightarrow f^{\otimes k}(v)
\end{equation}
can be expressed in its matrix representation, choosing $B$ as the basis for each copy of $\mathbb{R}^{n}$,
as the matrix $A(f) = (A_{I,J})$,
over all tuples $I \coloneqq (i_1, i_2, \dots, i_k), J \coloneqq (j_1, j_2, \dots, j_k) \in [n]^k$,
such that
\begin{equation} \label{cogredktransform}
	y_I = \sum_{J} A_{I,J} x_J		
\end{equation}
where
\begin{equation} \label{cogredmatrixprod}
	A_{I,J} = \prod_{r=1}^{k} a_{i_r,j_r}
\end{equation}
and
\begin{equation} \label{rhonbasisvecs}
	v = \sum_{J} x_Jb_J  \text{ and } f(v) = \sum_{J} y_Jb_J
\end{equation}
for some coefficients $x_J, y_J \in \mathbb{R}$
in the basis $\{b_J \mid J \in [n]^k\}$ of 
$(\mathbb{R}^{n})^{\otimes k}$,
where
\begin{equation} 
	b_J \coloneqq b_{j_1} \otimes b_{j_2} \otimes \dots \otimes b_{j_k} 
\end{equation}

As before, since $((\mathbb{R}^{n})^{\otimes k}, \rho_k)$ is a representation of $G$, we obtain the contragredient representation, $(((\mathbb{R}^{n})^{*})^{\otimes k}, (\rho_k^{-1})^\top)$
\begin{equation}
	(\rho_k^{-1})^\top : G \rightarrow GL(((\mathbb{R}^{n})^{*})^{\otimes k})
\end{equation}
where
\begin{equation}
	(\rho_k^{-1})^\top(f) : ((\mathbb{R}^{n})^{*})^{\otimes k}
	\rightarrow
	((\mathbb{R}^{n})^{*})^{\otimes k}
\end{equation}
is defined by
\begin{equation}
	(\rho_k^{-1})^\top(f)[u] : v \mapsto u(\rho_k^{-1}(f)(v))
\end{equation}
In particular, we see that $(\rho_k^{-1})^\top = ((\rho_1^{-1})^\top)^{\otimes k}$.

Hence, expressing $u \in 
	((\mathbb{R}^{n})^{*})^{\otimes k}$
in the dual basis $\{b_J^{*} \mid J \in [n]^k\}$ of 
$(\mathbb{R}^{n})^{\otimes k}$,
where
\begin{equation} 
	b_J^{*} \coloneqq b_{j_1}^{*} \otimes b_{j_2}^{*} \otimes \dots \otimes b_{j_k}^{*}
\end{equation}
as
\begin{equation} \label{contragredktensor}
	u = \sum_{J} p_Jb_J^{*} 
\end{equation}
and $(\rho_k^{-1})^\top(f)[u] \in 
	((\mathbb{R}^{n})^{*})^{\otimes k}$
as
\begin{equation}
	(\rho_k^{-1})^\top(f)[u] = \sum_{J} q_Jb_J^{*}
\end{equation}
we see that, for any $f \in G$, the linear transformation
\begin{equation}
	(\rho_k^{-1})^\top(f) : u \rightarrow (\rho_k^{-1})^\top(f)[u]
\end{equation}
can be expressed in its matrix representation form as
\begin{equation} \label{contragredmatrixsum}
	p_I = \sum_{J} q_JA_{J,I}
\end{equation}
where
\begin{equation} \label{contragredprod}
	A_{J,I} = \prod_{r=1}^{k} a_{j_r,i_r}
\end{equation}

\subsection{Brauer's Invariant Argument}

We adapt the argument given in \cite{Brauer} to construct an invariant of $G$ that is in bijective correspondence with an element of 
$\Hom_{G}((\mathbb{R}^{n})^{\otimes k}, (\mathbb{R}^{n})^{\otimes l})$.

A linear map $\phi : (\mathbb{R}^{n})^{\otimes k} \rightarrow
(\mathbb{R}^{n})^{\otimes l}$
is an element of 
$\Hom_{G}((\mathbb{R}^{n})^{\otimes k}, (\mathbb{R}^{n})^{\otimes l})$
if and only if
\begin{equation} \label{homequiv}
	\phi \circ \rho_k(f) = \rho_l(f) \circ \phi
\end{equation}
for all $f \in G$.

Choosing the basis $B$ as the basis for each copy of $\mathbb{R}^{n}$,
the matrix representation of (\ref{homequiv}) is
\begin{equation} \label{homequivmatrix}
	C(\phi)K(f) = L(f)C(\phi)
\end{equation}
where $K(f) = (K_{I,J})$ and $L(f) = (L_{I,J})$ are as in (\ref{kpowerrep})
and $C(\phi) = (C_{I,J})$.

Hence, (\ref{homequivmatrix}) gives
\begin{equation} \label{homequivequality}
	\sum_{J \in [n]^{k}} C_{I,J}K_{J,M} = \sum_{J \in [n]^{l}} L_{I,J}C_{J,M}
\end{equation}
where $I \in [n]^l$ and $M \in [n]^k$.

Brauer's trick is as follows.

Let $v(1), v(2), \dots, v(k)$ be elements of $\mathbb{R}^{n}$, and suppose that they are all mapped by the same transformation $\rho_1(f)$, for some $f \in G$.
Then, by (\ref{rho1basisvecs}),
in the basis $B$ of $\mathbb{R}^{n}$, we see that
\begin{equation}
	v(r) = \sum_{j} x_j(r)b_j
\end{equation}
for all $r \in [k]$, and so, by (\ref{rho1action}),
we have that
\begin{equation}
	y_i(r) = \sum_{j} a_{i,j} x_j(r)
\end{equation}
for all $r \in [k]$.

Then, considering the tensor product $v(1) \otimes v(2) \otimes \dots \otimes v(k)$, an element of 
$(\mathbb{R}^{n})^{\otimes k}$, 
and considering its transformation under $\rho_k(f)$, for the same $f \in G$, we see that the coefficient of the basis vector $b_{J}$ for $v(1) \otimes v(2) \otimes \dots \otimes v(k)$, as in (\ref{rhonbasisvecs}), is
\begin{equation}
	\prod_{r=1}^{k} x_{j_r}(r)
\end{equation}
and the coefficient of the basis vector $b_{I}$ for $\rho_k(f)[v(1) \otimes v(2) \otimes \dots \otimes v(k)]$ is
\begin{equation}
	\prod_{r=1}^{k} y_{i_r}(r)
\end{equation}
and that (\ref{cogredmatrixprod}) holds, namely that
\begin{equation}
	K_{I,J} = \prod_{r=1}^{k} a_{i_r,j_r}
\end{equation}
%(the relationship of $A_{I,J}$ to $a_{i_r,j_r}$).
Hence, by (\ref{cogredktransform}), we have that
\begin{equation} \label{cogredequality}
	\prod_{r=1}^{k} y_{i_r}(r)
	= \sum_{J \in [n]^k} K_{I,J} \prod_{r=1}^{k} x_{j_r}(r)
\end{equation}
Also, let $u(1), u(2), \dots, u(l)$ be elements of $(\mathbb{R}^{n})^{*}$, the dual space of $\mathbb{R}^{n}$.
Then, by (\ref{contragredbasisvec}), in the dual basis $B^{*}$, we have that
\begin{equation}
	u(t) = \sum_{j} p_j(t)b_j^{*}
\end{equation}
for all $t \in [l]$, and so, by (\ref{contragredsum}),
we see that
\begin{equation}
	p_i(t) = \sum_{j} q_j(t)a_{j,i}
\end{equation}
for all $t \in [l]$.

Then, considering the tensor product $u(1) \otimes u(2) \otimes \dots \otimes u(l)$, an element of $((\mathbb{R}^{n})^{*})^{\otimes l}$, 
and considering its transformation under $(\rho_l^{-1})^\top(f)$, for the same $f \in G$, we see that the coefficient of the basis vector $b_{I}^{*}$ for $u(1) \otimes u(2) \otimes \dots \otimes u(l)$, as in (\ref{contragredktensor}),
is
\begin{equation}
	\prod_{t=1}^{l} p_{i_t}(t)
\end{equation}
and the coefficient of the basis vector $b_{J}^{*}$ for $(\rho_l^{-1})^\top(f)[u(1) \otimes u(2) \otimes \dots \otimes u(l)]$ is
\begin{equation}
	\prod_{t=1}^{l} q_{j_t}(t)
\end{equation}
and that (\ref{contragredprod}) holds, namely that
\begin{equation}
	L_{J,I} = \prod_{t=1}^{l} a_{j_t,i_t}
\end{equation}
%(the relationship of $A_{J,I}$ to $a_{j_r,i_r}$).
Hence, by (\ref{contragredmatrixsum}),
we have that
\begin{equation} \label{contragredequality}
	\prod_{t=1}^{l} p_{i_t}(t)
	= \sum_{J \in [n]^l} \prod_{t=1}^{l} q_{j_t}(t) L_{J,I} 
\end{equation}

Multiplying both sides of (\ref{homequivequality})
by
\begin{equation}
	\prod_{t=1}^{l} q_{i_t}(t)
	\prod_{r=1}^{k} x_{m_r}(r)
\end{equation}
adding over all tuples $I \in [n]^l, M \in [n]^k$, and applying (\ref{cogredequality}) and (\ref{contragredequality}) gives us, on the LHS
\begin{equation}
	\sum_{I \in [n]^l, J \in [n]^k} C_{I,J}
	\prod_{t=1}^{l} q_{i_t}(t)
	\prod_{r=1}^{k} y_{j_r}(r)
\end{equation}
and on the RHS
\begin{equation}
	\sum_{J \in [n]^l, M \in [n]^k} C_{J,M} 
	\prod_{t=1}^{l} p_{j_t}(t)
	\prod_{r=1}^{k} x_{m_r}(r)
\end{equation}
This means that
\begin{equation} \label{Ginvariant}
	\sum_{I \in [n]^l, J \in [n]^k} C_{I,J} 
	\prod_{t=1}^{l} p_{i_t}(t)
	\prod_{r=1}^{k} x_{j_r}(r)
\end{equation}
is an invariant for the group $G$, that is, it is a linear transformation
\begin{equation}
	((\mathbb{R}^{n})^{*})^{\otimes l} 
	\otimes 
	(\mathbb{R}^{n})^{\otimes k} 
	\rightarrow \mathbb{R}
\end{equation}
which maps an element of the form
\begin{equation} \label{invarianttensorelement}
	u(1) \otimes u(2) \otimes \dots \otimes u(l)	
	\otimes 
	v(1) \otimes v(2) \otimes \dots \otimes v(k) 
\end{equation}
to (\ref{Ginvariant}) that is invariant under the action of $G$.

We see that each stage of the argument, from (\ref{homequiv}) to (\ref{invarianttensorelement}),
is an if and only if statement, since
any invariant of $G$ which is linear in any subset 
$\{v(1), v(2), \dots, v(k)\}$
of $k$ vectors of $\mathbb{R}^{n}$ and any subset 
$\{u(1), u(2), \dots, u(l)\}$
of $l$ vectors in $(\mathbb{R}^{n})^{*}$, and is a homogeneous function of their union, must be of the form 
(\ref{Ginvariant})
since any invariant of these $l+k$ elements must be a linear combination of the elements
$\prod_{t=1}^{l} p_{i_t}(t) \prod_{r=1}^{k} x_{j_r}(r)$
where $x_{j_r}(r)$ is the $j_r^\text{th}$ coefficient of $v(r)$ when expressed in some basis $B$ of $\mathbb{R}^{n}$, and
$p_{i_t}(t)$ is the $i_t^\text{th}$ coefficient of $u(t)$ when expressed in its dual basis $B^{*}$.
Hence, starting from
(\ref{Ginvariant}) 
and
running the argument in reverse 
gives (\ref{homequiv}), and therefore shows that it is an if and only if statement.

\begin{comment}
This means that the elements of 
	$\Hom_G((\mathbb{R}^{n})^{\otimes k}, (\mathbb{R}^{n})^{\otimes l})$
are characterised by the fact that the entries of their matrix representations, in any basis of $\mathbb{R}^{n}$,
appear as the coefficients of the invariants 
\begin{equation}
	((\mathbb{R}^{n})^{*})^{\otimes l} 
	\otimes 
	(\mathbb{R}^{n})^{\otimes k} 
	\rightarrow \mathbb{R}
\end{equation}
of $G$ of the form (\ref{Ginvariant}).

In particular, this means that the elements of 
	$\Hom_G((\mathbb{R}^{n})^{\otimes k}, (\mathbb{R}^{n})^{\otimes l})$, 
having chosen the basis $B$ for each copy of $\mathbb{R}^{n}$,
are spanned by a set of matrices, where the entries of each matrix in the set appear as the coefficients of an invariant 
\begin{equation}
	((\mathbb{R}^{n})^{*})^{\otimes l} 
	\otimes 
	(\mathbb{R}^{n})^{\otimes k} 
	\rightarrow \mathbb{R}
\end{equation}
of $G$ of the form (\ref{Ginvariant}).
\end{comment}

In particular, this means that each element of 
	$\Hom_G((\mathbb{R}^{n})^{\otimes k}, (\mathbb{R}^{n})^{\otimes l})$, 
having chosen the basis $B$ for each copy of $\mathbb{R}^{n}$,
is in bijective correspondence with an invariant
\begin{equation}
	((\mathbb{R}^{n})^{*})^{\otimes l} 
	\otimes 
	(\mathbb{R}^{n})^{\otimes k} 
	\rightarrow \mathbb{R}
\end{equation}
of $G$ of the form (\ref{Ginvariant}), as desired.

\section{First Fundamental Theorems for $O(n)$, $Sp(n)$ and $SO(n)$} \label{FFTfortheGroups}

We state, without proof, the First Fundamental Theorems for $O(n)$, $Sp(n)$ and $SO(n)$. See \cite{goodman} for more details.

\begin{theorem}[First Fundamental Theorem for $O(n)$]
	Let $n \in \mathbb{Z}_{\geq 1}$, and suppose that 
	the real vector space $\mathbb{R}^{n}$ has associated with it
	a non-degenerate, symmetric, bilinear form
	$(\cdot{,}\cdot)$,
	as in Section~\ref{groupsOnSOnSpn}.

	Let us choose the standard basis for $\mathbb{R}^{n}$, so that
	$(\cdot{,}\cdot)$ becomes the Euclidean inner product on $\mathbb{R}^{n}$, as defined in (\ref{euclideaninnprod}).

	If $f : (\mathbb{R}^{n})^{\otimes (l+k)} \rightarrow \mathbb{R}$ is a polynomial function on elements in $(\mathbb{R}^{n})^{\otimes (l+k)}$
	of the form
	\begin{equation}
		u(1) \otimes u(2) \otimes \dots \otimes u(l) \otimes v(1) \otimes v(2) \otimes \dots \otimes v(k)	
	\end{equation}
	that is $O(n)$-invariant, then it must be a polynomial of the Euclidean inner products
	\begin{equation}
		(u(i), u(j)), (u(i), v(j)), (v(i), v(j)) 
	\end{equation}
\end{theorem}

\begin{theorem}[First Fundamental Theorem for $Sp(n)$]
	Let $n \in \mathbb{Z}_{\geq 2}$ be even, and suppose that
	the real vector space $\mathbb{R}^{n}$ has associated with it
	a non-degenerate, skew-symmetric, bilinear form
	$\langle\cdot{,}\cdot\rangle$,
	as in Section~\ref{groupsOnSOnSpn}.

	Let us choose the symplectic basis for $\mathbb{R}^{n}$, so that
	$\langle\cdot{,}\cdot\rangle$ becomes the form given in 
	(\ref{skewsymmform}).

	Note that, in this basis,
	the non-degenerate, \textit{symmetric}, bilinear form
	$(\cdot{,}\cdot)$ 
	which we can also associate with $\mathbb{R}^{n}$,
	becomes the Euclidean inner product on $\mathbb{R}^{n}$, as defined in (\ref{euclideaninnprod}), since the symplectic basis is standard with respect to $(\cdot{,}\cdot)$.
	If $f : (\mathbb{R}^{n})^{\otimes (l+k)} \rightarrow \mathbb{R}$ is a polynomial function on elements in $(\mathbb{R}^{n})^{\otimes (l+k)}$
	of the form
	\begin{equation}
		u(1) \otimes u(2) \otimes \dots \otimes u(l) \otimes v(1) \otimes v(2) \otimes \dots \otimes v(k)	
	\end{equation}
	that is $Sp(n)$-invariant, then it must be a polynomial of the Euclidean inner products
	\begin{equation}
		(u(i), v(j)) 
	\end{equation}
	together with the skew products
	\begin{equation} \label{Sp(n)skew}
		\langle u(i), u(j) \rangle, \langle v(i), v(j) \rangle
	\end{equation}
	such that $i < j$ in (\ref{Sp(n)skew}).
\end{theorem}

\begin{theorem}[First Fundamental Theorem for $SO(n)$]
	Let $n \in \mathbb{Z}_{\geq 1}$, and suppose that 
	the real vector space $\mathbb{R}^{n}$ has associated with it
	a non-degenerate, symmetric, bilinear form
	$(\cdot{,}\cdot)$,
	as in Section~\ref{groupsOnSOnSpn}.

	Let us choose the standard basis for $\mathbb{R}^{n}$, so that
	$(\cdot{,}\cdot)$ becomes the Euclidean inner product on $\mathbb{R}^{n}$, as defined in (\ref{euclideaninnprod}).

	If $f : (\mathbb{R}^{n})^{\otimes (l+k)} \rightarrow \mathbb{R}$ is a polynomial function on elements in $(\mathbb{R}^{n})^{\otimes (l+k)}$
	of the form
	\begin{equation}
		u(1) \otimes u(2) \otimes \dots \otimes u(l) \otimes v(1) \otimes v(2) \otimes \dots \otimes v(k)	
	\end{equation}
	that is $SO(n)$-invariant, then it must be a polynomial of the Euclidean inner products
	\begin{equation}
		(u(i), u(j)), (u(i), v(j)), (v(i), v(j)) 
	\end{equation}
	together with the $n \times n$ subdeterminants of the $n \times (l+k)$ matrix $M$, which is the matrix having as its columns:
	\begin{equation}
		M \coloneqq 
		\begin{pmatrix}
			\mid & \mid &  & \mid & \mid & \mid & & \mid \\
			u(1) & u(2) & \dots & u(l) & v(1) & v(2) & \dots & v(k) \\
			\mid & \mid & & \mid & \mid & \mid & & \mid 
		\end{pmatrix}
	\end{equation}
\end{theorem}

\section{Bijective Correspondence between the Brauer and Brauer--Grood vector spaces and the Invariants for $O(n)$, $Sp(n)$ and $SO(n)$}

We now provide a proof of the results given in Theorems
\ref{spanningsetO(n)}, \ref{spanningsetSp(n)}, and \ref{spanningsetSO(n)}.

It can be shown that $\mathbb{R}^{n}$, as a representation of each of the groups $G = O(n)$, $Sp(n)$ and $SO(n)$ (for $Sp(n)$, $n = 2m$), is isomorphic to its dual space $(\mathbb{R}^{n})^{*}$, by using the appropriate bilinear form that is used to define each of the groups in question. 
See \cite{goodman} for more details.

Hence, for each group $G$, we can apply its First Fundamental Theorem to the invariant given in (\ref{Ginvariant}), now considered as a function
$(\mathbb{R}^{n})^{\otimes (l+k)} \rightarrow \mathbb{R}$, noting that 
	each term of the polynomial (\ref{Ginvariant}) contains each of the vectors
$u(1), u(2), \dots, u(l), v(1), v(2), \dots, v(k)$ exactly once. 

Consequently, we obtain a spanning set of invariants
	$(\mathbb{R}^{n})^{\otimes (l+k)} \rightarrow \mathbb{R}$
	for each of 
	$O(n)$, $Sp(n)$ and $SO(n)$ (for $Sp(n)$, $n = 2m$).

\begin{theorem}[Spanning Set of Invariants
	$(\mathbb{R}^{n})^{\otimes (l+k)} \rightarrow \mathbb{R}$
	for $O(n)$] 
	\label{spannsetO(n)}

	%If $l+k$ is even, then 
	The functions 
	\begin{equation} \label{Oninvariantsform}
			(z(1), z(2))
			(z(3), z(4))
			\dots	
			(z(l+k-1), z(l+k))
	\end{equation}
	where		
	$z(1), \dots, z(l+k)$
	is a permutation of 
	$u(1), u(2), \dots, u(l), v(1), v(2), \dots, v(k)$
	form a spanning set of invariants
	$(\mathbb{R}^{n})^{\otimes (l+k)} \rightarrow \mathbb{R}$
	for $O(n)$.
	
	Clearly, functions of the form (\ref{Oninvariantsform}) can only be formed when $l+k$ is even, hence
	there are no invariants 
	$(\mathbb{R}^{n})^{\otimes (l+k)} \rightarrow \mathbb{R}$
	for $O(n)$ when $l+k$ is odd.
\end{theorem}

%\begin{proof}
	%Apply the First Fundamental Theorem for $O(n)$ to the invariant given in (\ref{Ginvariant}), noting that 
	%each term of the polynomial (\ref{Ginvariant}) contains each of the vectors
%$u(1), u(2), \dots, u(l), v(1), v(2), \dots, v(k)$ exactly once. 
%\end{proof}

\begin{theorem}[Spanning Set of Invariants
	$(\mathbb{R}^{n})^{\otimes (l+k)} \rightarrow \mathbb{R}$
	for $Sp(n)$, $n = 2m$]
	\label{spannsetSp(n)}

	The functions 
	\begin{equation} \label{Spninvariantsform}
			[z(1), z(2)]
			[z(3), z(4)]
			\dots	
			[z(l+k-1), z(l+k)]
	\end{equation}
	where		
	$z(1), \dots, z(l+k)$
	is a permutation of 
	$u(1), u(2), \dots, u(l), v(1), v(2), \dots, v(k)$
	and
	\begin{equation}
		[z(i), z(i+1)]	
		\coloneqq
		\begin{cases}
			(z(i), z(i+1))	& 
			\parbox{6cm}{if 
			$z(i) = u(j)$ and $z(i+1) = v(m)$, or 
			$z(i) = v(m)$ and $z(i+1) = u(j)$,
			for some $j \in [l]$, $m \in [k]$} \\
			\\
			\langle z(i), z(i+1)\rangle
			& \text{otherwise.}
    		\end{cases}
	\end{equation}
	form a spanning set of invariants
	$(\mathbb{R}^{n})^{\otimes (l+k)} \rightarrow \mathbb{R}$
	for $Sp(n)$, with $n = 2m$.

	Clearly, functions of the form (\ref{Spninvariantsform}) can only be formed when $l+k$ is even, hence
	there are no invariants 
	$(\mathbb{R}^{n})^{\otimes (l+k)} \rightarrow \mathbb{R}$
	for $Sp(n)$ when $l+k$ is odd.
\end{theorem}

\begin{theorem}[Spanning Set of Invariants
	$(\mathbb{R}^{n})^{\otimes (l+k)} \rightarrow \mathbb{R}$
	for $SO(n)$]
	\label{spannsetSO(n)}

	Functions of the form (\ref{Oninvariantsform}) together with functions of the form
	\begin{equation} \label{SOninvariantsform}
		\det(z(1), \dots, z(n)) 
			(z(n+1), z(n+2))
			\dots	
			(z(l+k-1), z(l+k))
	\end{equation}
	where		
	$z(1), \dots, z(l+k)$
	is a permutation of 
	$u(1), u(2), \dots, u(l), v(1), v(2), \dots, v(k)$
	form a spanning set of invariants
	$(\mathbb{R}^{n})^{\otimes (l+k)} \rightarrow \mathbb{R}$
	for $SO(n)$.

	Clearly, functions of the form (\ref{SOninvariantsform}) can only be formed when $n \leq l+k$, and 
	either when $n$ is odd and $l+k$ is odd, 
	or when $n$ is even and $l+k$ is even.
\end{theorem}

%The critical step is that
We can now construct a bijective correspondence between
each of the functions 
(\ref{Oninvariantsform}),
(\ref{Spninvariantsform}), and
(\ref{SOninvariantsform}),
and either a $(k,l)$--Brauer diagram or an $(l+k)\backslash n$--diagram, as follows.

Indeed, consider the spanning set of invariants of the form
(\ref{Oninvariantsform}).
Then we can associate a $(k,l)$--Brauer diagram with each element of the set by labelling the top $l$ vertices by 
$u(1), u(2), \dots, u(l)$
and the bottom $k$ vertices by $v(1), v(2), \dots, v(k)$, and drawing an edge between two vertices if and only if they appear in the same inner product in (\ref{Oninvariantsform}).

We do a similar thing for the spanning set of invariants of the form
(\ref{Spninvariantsform}), associating a $(k,l)$--Brauer diagram with
each element of the set, labelling the vertices in the same mamner, and drawing an edge between two vertices if and only if they appear in the same inner or skew product in (\ref{Spninvariantsform}).

Finally, consider the functions of the form 
(\ref{SOninvariantsform}).
Then we can associate 
an $(l+k)\backslash n$--diagram
with each element of the set 
by labelling the top $l$ vertices by 
$u(1), u(2), \dots, u(l)$
and the bottom $k$ vertices by $v(1), v(2), \dots, v(k)$, leaving the vertices $z(1), \dots, z(n)$ free and drawing an edge between the other vertices if and only if they appear in the same inner product in
(\ref{SOninvariantsform}).

As a result, we have constructed a bijective correspondence between a spanning set of invariants 
$(\mathbb{R}^{n})^{\otimes (l+k)} \rightarrow \mathbb{R}$
for $O(n)$ with 
the set of $(k,l)$--Brauer diagrams whose span is
the Brauer vector space $B_k^l(n)$,
for $SO(n)$ with 
the set of $(k,l)$--Brauer diagrams together with the set of $(l+k)\backslash n$--diagrams whose span is
the Brauer--Grood vector space $D_k^l(n)$, and
for $Sp(n)$ ($n = 2m$) with 
the set of $(k,l)$--Brauer diagrams whose span is
the Brauer vector space $B_k^l(n)$.

Consequently, for each group $G$, the bijective correspondence 
(\ref{Ginvariant})
that exists
between the spannning set of invariants 
$(\mathbb{R}^{n})^{\otimes (l+k)} \rightarrow \mathbb{R}$
(given in Theorems 
	\ref{spannsetO(n)}, 
	\ref{spannsetSp(n)}, and
	\ref{spannsetSO(n)})
and a spanning set of matrices for 
	$\Hom_G((\mathbb{R}^{n})^{\otimes k}, (\mathbb{R}^{n})^{\otimes l})$
in the standard/symplectic basis of $\mathbb{R}^{n}$, 
and the bijective correspondence 
that exists
between the spannning set of invariants 
$(\mathbb{R}^{n})^{\otimes (l+k)} \rightarrow \mathbb{R}$
and 
the set of diagrams that span
either the Brauer vector space
$B_k^l(n)$, for $O(n)$ and $Sp(n)$, or the Brauer--Grood vector space
$D_k^l(n)$, for $SO(n)$, together prove the results given in
Theorems 
\ref{spanningsetO(n)}, \ref{spanningsetSp(n)}, and \ref{spanningsetSO(n)}.

\section{Examples} \label{examples}

In the following, in order to save space, we use some temporary notation to denote a sum of a number of weight parameters.
We represent a sum of weight parameters, where the sum is over some index set $A$, by a single element that is indexed by the set of indices itself, that is
\begin{equation}
	\lambda_A \coloneqq \sum_{i \in A} \lambda_i
\end{equation}
For example, $\lambda_{8,11,12}$ represents the sum $\lambda_{8} + \lambda_{11} + \lambda_{12}$.

\subsection{$O(n)$}

%\underline{1. A Basis of $\End_{O(2)}((\mathbb{R}^{2})^{\otimes 2})$}
\subsubsection{A Basis of $\End_{O(2)}((\mathbb{R}^{2})^{\otimes 2})$}

We consider the surjective map
\begin{equation}
	\Phi_{2,2}^2 : B_2^2(2) \rightarrow \End_{O(2)}((\mathbb{R}^{2})^{\otimes 2})
\end{equation}
and apply Theorem \ref{spanningsetO(n)},
noting that $l + k$ is even.
Also, as $2n \geq l + k$, this map is an isomorphism of vector spaces, hence the images of the basis diagrams of $B_2^2(2)$ forms a basis of 
$\End_{O(2)}((\mathbb{R}^{2})^{\otimes 2})$.

Figure \ref{matrix2,2} shows how to find the basis of
$\End_{O(2)}((\mathbb{R}^{2})^{\otimes 2})$
from the basis of $B_2^2(2)$.

\begin{figure}[ht]
	\begin{center}
\begin{tblr}{
  colspec = {X[c,h]X[c]X[c]},
  stretch = 0,
  rowsep = 6pt,
  hlines = {1pt},
  vlines = {1pt},
}
	{Basis Diagram $d_\beta$} 	& {Matrix Entries}	& 
	{Basis Element of 
	$\End_{O(2)}((\mathbb{R}^{2})^{\otimes 2})$}\\
	\scalebox{0.6}{\begin{tikzpicture}
	\begin{pgfonlayer}{nodelayer}
		\node [style=none] (0) at (0, 1) {$1$};
		\node [style=node] (1) at (0, 0) {};
		\node [style=node] (2) at (0, -3) {};
		\node [style=node] (3) at (3, 0) {};
		\node [style=node] (4) at (3, -3) {};
		\node [style=none] (5) at (3, 1) {$2$};
		\node [style=none] (6) at (0, -4) {$3$};
		\node [style=none] (7) at (3, -4) {$4$};
		\node [style=none] (8) at (-0.5, 0.5) {};
		\node [style=none] (9) at (-0.5, -3.5) {};
		\node [style=none] (10) at (3.5, -3.5) {};
		\node [style=none] (11) at (3.5, 0.5) {};
	\end{pgfonlayer}
	\begin{pgfonlayer}{edgelayer}
		\draw [style={dash_2}] (8.center)
			 to (9.center)
			 to (10.center)
			 to (11.center)
			 to cycle;
		\draw [style=thick] (2) to (4);
		\draw [style=thick] (3) to (1);
	\end{pgfonlayer}
\end{tikzpicture}} & $(\delta_{i_1, i_2}\delta_{j_1,j_2})$
	& 
	\scalebox{0.75}{
	$
	\NiceMatrixOptions{code-for-first-row = \scriptstyle \color{blue},
                   	   code-for-first-col = \scriptstyle \color{blue}
	}
	\begin{bNiceArray}{*{2}{c}|*{2}{c}}[first-row,first-col]
				& 1,1 	& 1,2	& 2,1	& 2,2 	\\
		1,1		& 1	& 0	& 0	& 1	\\
		1,2		& 0	& 0	& 0	& 0	\\
		\cline{1-4}
		2,1		& 0	& 0	& 0	& 0	\\
		2,2		& 1	& 0	& 0	& 1
	\end{bNiceArray}
	$}
	\\
	\scalebox{0.6}{\begin{tikzpicture}
	\begin{pgfonlayer}{nodelayer}
		\node [style=none] (0) at (0, 1) {$1$};
		\node [style=node] (1) at (0, 0) {};
		\node [style=node] (2) at (0, -3) {};
		\node [style=node] (3) at (3, 0) {};
		\node [style=node] (4) at (3, -3) {};
		\node [style=none] (5) at (3, 1) {$2$};
		\node [style=none] (6) at (0, -4) {$3$};
		\node [style=none] (7) at (3, -4) {$4$};
		\node [style=none] (8) at (-0.5, 0.5) {};
		\node [style=none] (9) at (-0.5, -3.5) {};
		\node [style=none] (10) at (3.5, -3.5) {};
		\node [style=none] (11) at (3.5, 0.5) {};
	\end{pgfonlayer}
	\begin{pgfonlayer}{edgelayer}
		\draw [style={dash_2}] (8.center)
			 to (9.center)
			 to (10.center)
			 to (11.center)
			 to cycle;
		\draw [style=thick] (2) to (1);
		\draw [style=thick] (4) to (3);
	\end{pgfonlayer}
\end{tikzpicture}}	& 
	$(\delta_{i_1, j_1}\delta_{i_2,j_2})$
	& 
	\scalebox{0.75}{
	$
	\NiceMatrixOptions{code-for-first-row = \scriptstyle \color{blue},
                   	   code-for-first-col = \scriptstyle \color{blue}
	}
	\begin{bNiceArray}{*{2}{c}|*{2}{c}}[first-row,first-col]
				& 1,1 	& 1,2	& 2,1	& 2,2 	\\
		1,1		& 1	& 0	& 0	& 0	\\
		1,2		& 0	& 1	& 0	& 0	\\
		%\hline
		\cline{1-4}
		2,1		& 0	& 0	& 1	& 0	\\
		2,2		& 0	& 0	& 0	& 1
	\end{bNiceArray}
	$}
	\\
	\scalebox{0.6}{\begin{tikzpicture}
	\begin{pgfonlayer}{nodelayer}
		\node [style=none] (0) at (0, 1) {$1$};
		\node [style=node] (1) at (0, 0) {};
		\node [style=node] (2) at (0, -3) {};
		\node [style=node] (3) at (3, 0) {};
		\node [style=node] (4) at (3, -3) {};
		\node [style=none] (5) at (3, 1) {$2$};
		\node [style=none] (6) at (0, -4) {$3$};
		\node [style=none] (7) at (3, -4) {$4$};
		\node [style=none] (8) at (-0.5, 0.5) {};
		\node [style=none] (9) at (-0.5, -3.5) {};
		\node [style=none] (10) at (3.5, -3.5) {};
		\node [style=none] (11) at (3.5, 0.5) {};
	\end{pgfonlayer}
	\begin{pgfonlayer}{edgelayer}
		\draw [style={dash_2}] (8.center)
			 to (9.center)
			 to (10.center)
			 to (11.center)
			 to cycle;
		\draw [style=thick] (1) to (4);
		\draw [style=thick] (2) to (3);
	\end{pgfonlayer}
\end{tikzpicture}}	& 
	$(\delta_{i_1, j_2}\delta_{i_2,j_1})$
	& 
	\scalebox{0.75}{
	$
	\NiceMatrixOptions{code-for-first-row = \scriptstyle \color{blue},
                   	   code-for-first-col = \scriptstyle \color{blue}
	}
	\begin{bNiceArray}{*{2}{c}|*{2}{c}}[first-row,first-col]
				& 1,1 	& 1,2	& 2,1	& 2,2 	\\
		1,1		& 1	& 0	& 0	& 0	\\
		1,2		& 0	& 0	& 1	& 0	\\
		%\hline
		\cline{1-4}
		2,1		& 0	& 1	& 0	& 0	\\
		2,2		& 0	& 0	& 0	& 1
	\end{bNiceArray}
	$}
	\\
\end{tblr}
		\caption{The images under $\Phi_{2,2}^2$ of the basis diagrams of $B_2^2(2)$ make up a basis of $\End_{O(2)}((\mathbb{R}^{2})^{\otimes 2})$.}
			%A table showing the basis diagrams of $B_2^2(2)$ whose images under $\Phi_{2,2}^2$ make up a basis of $\End_{O(2)}((\mathbb{R}^{2})^{\otimes 2})$.}
  	\label{matrix2,2}
	\end{center}
\end{figure}

Hence, any element of 
$\End_{O(2)}((\mathbb{R}^{2})^{\otimes 2})$,
in the basis of matrix units of 
$\End((\mathbb{R}^{2})^{\otimes 2})$,
is of the form
%\begin{figure}[ht]
\begin{equation}
	\NiceMatrixOptions{code-for-first-row = \scriptstyle \color{blue},
                   	   code-for-first-col = \scriptstyle \color{blue}
	}
	\renewcommand{\arraystretch}{1.5}
	\begin{bNiceArray}{*{2}{c}|*{2}{c}}[first-row,first-col]
				& 1,1 	& 1,2	& 2,1	& 2,2 	\\
		1,1		& \lambda_{1,2,3}	& 0	& 0	& \lambda_1	\\
		1,2		& 0	& \lambda_2 	& \lambda_3	& 0	\\
		\hline
		2,1		& 0	& \lambda_3	& \lambda_2	& 0	\\
		2,2		& \lambda_1	& 0	& 0	& \lambda_{1,2,3}
	\end{bNiceArray}
\end{equation}
%\end{figure}
for scalars $\lambda_1, \lambda_2, \lambda_3 \in \mathbb{R}$.

%\newpage

%\underline{2. A Basis of $\End_{O(3)}((\mathbb{R}^{3})^{\otimes 3})$}
\subsubsection{A Basis of $\End_{O(3)}((\mathbb{R}^{3})^{\otimes 3})$}

We consider the surjective map
\begin{equation}
	\Phi_{3,3}^3 : B_3^3(3) \rightarrow \End_{O(3)}((\mathbb{R}^{3})^{\otimes 3})
\end{equation}
and apply Theorem \ref{spanningsetO(n)},
noting that $l + k$ is even.
Also, as $2n \geq l + k$, this map is an isomorphism of vector spaces, hence the images of the basis diagrams of $B_3^3(3)$ forms a basis of 
$\End_{O(3)}((\mathbb{R}^{3})^{\otimes 3})$.

The basis diagrams of $B_3^3(3)$ are
\begin{center}
	\scalebox{0.45}{\input{brauer33.tikz}}
\end{center}
Taking their images under $\Phi_{3,3}^3$, we see that
any element of 
$\End_{O(3)}((\mathbb{R}^{3})^{\otimes 3})$,
in the basis of matrix units of 
$\End((\mathbb{R}^{3})^{\otimes 3})$,
is of the form
\begin{equation} \label{EndO3R3tensor3}
	\scalebox{0.465}{$
	\NiceMatrixOptions{code-for-first-row = \scriptstyle \color{blue},
                   	   code-for-first-col = \scriptstyle \color{blue}
	}
	\renewcommand{\arraystretch}{1.5}
	\begin{bNiceArray}{*{3}{c}|*{3}{c}|*{3}{c}|*{3}{c}|*{3}{c}|*{3}{c}|*{3}{c}|*{3}{c}|*{3}{c}}[first-row,first-col]
		\RowStyle[cell-space-limits=3pt]{\rotate}
				& 1,1,1 	& 1,1,2		& 1,1,3 	
				& 1,2,1 	& 1,2,2		& 1,2,3 	
				& 1,3,1 	& 1,3,2		& 1,3,3 	
				& 2,1,1 	& 2,1,2		& 2,1,3 	
				& 2,2,1 	& 2,2,2		& 2,2,3 	
				& 2,3,1 	& 2,3,2		& 2,3,3 	
				& 3,1,1 	& 3,1,2		& 3,1,3 	
				& 3,2,1 	& 3,2,2		& 3,2,3 	
				& 3,3,1 	& 3,3,2		& 3,3,3 \\
		1,1,1		& \lambda_{1,\dots,15} & 0	& 0 
				& 0 & \lambda_{8,11,12}	& 0 
				& 0 & 0	& \lambda_{8,11,12} 
				& 0 & \lambda_{9,14,15}	& 0 
				& \lambda_{7,10,13} & 0	& 0 
				& 0 & 0	& 0 
				& 0 & 0	& \lambda_{9,14,15} 
				& 0 & 0	& 0 
				& \lambda_{7,10,13} & 0	& 0 \\
		1,1,2		& 0 & \lambda_{1,2,7}	& 0 
				& \lambda_{4,6,15} & 0	& 0 
				& 0 & 0	& 0 
				& \lambda_{3,5,11} & 0	& 0 
				& 0 & \lambda_{7,11,15}	& 0 
				& 0 & 0	& \lambda_{11} 
				& 0 & 0	& 0 
				& 0 & 0	& \lambda_{15} 
				& 0 & \lambda_{7} & 0 \\
		1,1,3		& 0 & 0	& \lambda_{1,2,7} 
				& 0 & 0	& 0 
				& \lambda_{4,6,15} & 0	& 0 
				& 0 & 0	& 0 
				& 0 & 0	& \lambda_{7} 
				& 0 & \lambda_{15}	& 0 
				& \lambda_{3,5,11} & 0	& 0 
				& 0 & \lambda_{11}	& 0 
				& 0 & 0	& \lambda_{7,11,15} \\
		\hline
		1,2,1		& 0 & \lambda_{3,4,13}	& 0 
				& \lambda_{1,5,9} & 0	& 0 
				& 0 & 0	& 0 
				& \lambda_{2,6,12} & 0	& 0 
				& 0 & \lambda_{9,12,13}	& 0 
				& 0 & 0	& \lambda_{12} 
				& 0 & 0	& 0 
				& 0 & 0	& \lambda_{9} 
				& 0 & \lambda_{13} & 0 \\
		1,2,2		& \lambda_{8,10,14} & 0	& 0 
				& 0 & \lambda_{1,4,8} & 0 
				& 0 & 0	& \lambda_{8} 
				& 0 & \lambda_{2,3,14} & 0 
				& \lambda_{5,6,10} & 0	& 0 
				& 0 & 0	& 0 
				& 0 & 0	& \lambda_{14} 
				& 0 & 0	& 0 
				& \lambda_{10} & 0 & 0 \\
		1,2,3		& 0 & 0	& 0 
				& 0 & 0	& \lambda_1 
				& 0 & \lambda_4	& 0 
				& 0 & 0	& \lambda_2 
				& 0 & 0	& 0 
				& \lambda_6 & 0	& 0 
				& 0 & \lambda_3	& 0 
				& \lambda_5 & 0	& 0 
				& 0 & 0	& 0 \\
		\hline
		1,3,1		& 0 & 0	& \lambda_{3,4,13}
				& 0 & 0	& 0 
				& \lambda_{1,5,9} & 0	& 0 
				& 0 & 0	& 0 
				& 0 & 0	& \lambda_{13} 
				& 0 & \lambda_{9} & 0 
				& \lambda_{2,6,12} & 0	& 0 
				& 0 & \lambda_{12} & 0 
				& 0 & 0	& \lambda_{9,12,13} \\
		1,3,2		& 0 & 0	& 0 
				& 0 & 0	& \lambda_4
				& 0 & \lambda_1 & 0 
				& 0 & 0	& \lambda_3
				& 0 & 0	& 0 
				& \lambda_5 & 0	& 0 
				& 0 & \lambda_2	& 0 
				& \lambda_6 & 0	& 0 
				& 0 & 0	& 0 \\
		1,3,3		& \lambda_{8,10,14} & 0	& 0 
				& 0 & \lambda_{8} & 0 
				& 0 & 0	& \lambda_{1,4,8} 
				& 0 & \lambda_{14} & 0 
				& \lambda_{10} & 0 & 0 
				& 0 & 0	& 0 
				& 0 & 0	& \lambda_{2,3,14}
				& 0 & 0	& 0 
				& \lambda_{5,6,10} & 0 & 0 \\
		\hline
		2,1,1		& 0 & \lambda_{5,6,10}	& 0 
				& \lambda_{2,3,14} & 0	& 0 
				& 0 & 0	& 0 
				& \lambda_{1,4,8} & 0	& 0 
				& 0 & \lambda_{8,10,14}	& 0 
				& 0 & 0	& \lambda_{8} 
				& 0 & 0	& 0 
				& 0 & 0	& \lambda_{14} 
				& 0 & \lambda_{10} & 0 \\
		2,1,2		& \lambda_{9,12,13} & 0	& 0 
				& 0 & \lambda_{2,6,12}	& 0 
				& 0 & 0	& \lambda_{12} 
				& 0 & \lambda_{1,5,9}	& 0 
				& \lambda_{3,4,13} & 0	& 0 
				& 0 & 0	& 0 
				& 0 & 0	& \lambda_{9} 
				& 0 & 0	& 0 
				& \lambda_{13} & 0	& 0 \\
		2,1,3		& 0 & 0	& 0 
				& 0 & 0	& \lambda_2 
				& 0 & \lambda_6	& 0 
				& 0 & 0	& \lambda_1 
				& 0 & 0	& 0 
				& \lambda_4 & 0	& 0 
				& 0 & \lambda_5	& 0 
				& \lambda_3 & 0	& 0 
				& 0 & 0	& 0 \\
		\hline
		2,2,1		& \lambda_{7,11,15} & 0	& 0 
				& 0 & \lambda_{3,5,11}	& 0 
				& 0 & 0	& \lambda_{11} 
				& 0 & \lambda_{4,6,15}	& 0 
				& \lambda_{1,2,7} & 0	& 0 
				& 0 & 0	& 0 
				& 0 & 0	& \lambda_{15} 
				& 0 & 0	& 0 
				& \lambda_{7} & 0	& 0 \\
		2,2,2		& 0 & \lambda_{7,10,13}	& 0 
				& \lambda_{9,14,15} & 0	& 0 
				& 0 & 0	& 0 
				& \lambda_{8,11,12} & 0	& 0 
				& 0 & \lambda_{1,\dots,15} & 0 
				& 0 & 0	& \lambda_{8,11,12} 
				& 0 & 0	& 0 
				& 0 & 0	& \lambda_{9,14,15} 
				& 0 & \lambda_{7,10,13}	& 0 \\
		2,2,3		& 0 & 0	& \lambda_{7}
				& 0 & 0	& 0 
				& \lambda_{15} & 0 & 0 
				& 0 & 0	& 0 
				& 0 & 0	& \lambda_{1,2,7}
				& 0 & \lambda_{4,6,15}	& 0 
				& \lambda_{11} & 0	& 0 
				& 0 & \lambda_{3,5,11}	& 0 
				& 0 & 0	& \lambda_{7,11,15} \\
		\hline
		2,3,1		& 0 & 0	& 0 
				& 0 & 0	& \lambda_{3}
				& 0 & \lambda_{5} & 0 
				& 0 & 0	& \lambda_{4} 
				& 0 & 0	& 0 
				& \lambda_1 & 0	& 0 
				& 0 & \lambda_6	& 0 
				& \lambda_2 & 0	& 0 
				& 0 & 0	& 0 \\
		2,3,2		& 0 & 0	& \lambda_{13}
				& 0 & 0	& 0
				& \lambda_9 & 0 & 0 
				& 0 & 0 & 0
				& 0 & 0	& \lambda_{3,4,13}
				& 0 & \lambda_{1,5,9}	& 0 
				& \lambda_{12} & 0 & 0 
				& 0 & \lambda_{2,6,12}	& 0
				& 0 & 0	& \lambda_{9,12,13}\\
		2,3,3		& 0 & \lambda_{10} & 0 
				& \lambda_{14} & 0 & 0 
				& 0 & 0	& 0 
				& \lambda_{8} & 0	& 0 
				& 0 & \lambda_{8,10,14} & 0 
				& 0 & 0	& \lambda_{1,4,8} 
				& 0 & 0	& 0 
				& 0 & 0	& \lambda_{2,3,14} 
				& 0 & \lambda_{5,6,10}	& 0 \\
		\hline
		3,1,1		& 0 & 0	& \lambda_{5,6,10}
				& 0 & 0	& 0 
				& \lambda_{2,3,14} & 0	& 0 
				& 0 & 0	& 0 
				& 0 & 0	& \lambda_{10} 
				& 0 & \lambda_{14}	& 0 
				& \lambda_{1,4,8} & 0	& 0 
				& 0 & \lambda_{8} & 0 
				& 0 & 0	& \lambda_{8,10,14} \\
		3,1,2		& 0 & 0	& 0 
				& 0 & 0	& \lambda_6 
				& 0 & \lambda_2	& 0 
				& 0 & 0	& \lambda_5 
				& 0 & 0	& 0 
				& \lambda_3 & 0	& 0 
				& 0 & \lambda_1	& 0 
				& \lambda_4 & 0	& 0 
				& 0 & 0	& 0 \\
		3,1,3		& \lambda_{9,12,13} & 0	& 0 
				& 0 & \lambda_{12}	& 0 
				& 0 & 0	& \lambda_{2,6,12}
				& 0 & \lambda_{9} & 0 
				& \lambda_{13} & 0 & 0 
				& 0 & 0	& 0 
				& 0 & 0	& \lambda_{1,5,9}
				& 0 & 0	& 0 
				& \lambda_{3,4,13} & 0	& 0 \\
		\hline
		3,2,1		& 0 & 0	& 0 
				& 0 & 0	& \lambda_{5}
				& 0 & \lambda_{3} & 0 
				& 0 & 0	& \lambda_{6} 
				& 0 & 0	& 0 
				& \lambda_2 & 0	& 0 
				& 0 & \lambda_4	& 0 
				& \lambda_1 & 0	& 0 
				& 0 & 0	& 0 \\
		3,2,2		& 0 & 0	& \lambda_{10}
				& 0 & 0	& 0 
				& \lambda_{14} & 0 & 0 
				& 0 & 0	& 0 
				& 0 & 0	& \lambda_{5,6,10} 
				& 0 & \lambda_{2,3,14} & 0 
				& \lambda_{8} & 0 & 0 
				& 0 & \lambda_{1,4,8}	& 0 
				& 0 & 0	& \lambda_{8,10,14} \\
		3,2,3		& 0 & \lambda_{13} & 0 
				& \lambda_{9} & 0 & 0 
				& 0 & 0	& 0 
				& \lambda_{12} & 0	& 0 
				& 0 & \lambda_{9,12,13} & 0 
				& 0 & 0	& \lambda_{2,6,12}
				& 0 & 0	& 0 
				& 0 & 0	& \lambda_{1,5,9} 
				& 0 & \lambda_{3,4,13} & 0 \\
		\hline
		3,3,1		& \lambda_{7,11,15} & 0	& 0 
				& 0 & \lambda_{11} & 0 
				& 0 & 0	& \lambda_{3,5,11} 
				& 0 & \lambda_{15} & 0 
				& \lambda_{7} & 0 & 0 
				& 0 & 0	& 0 
				& 0 & 0	& \lambda_{4,6,15} 
				& 0 & 0	& 0 
				& \lambda_{1,2,7} & 0 & 0 \\
		3,3,2		& 0 & \lambda_{7} & 0 
				& \lambda_{15} & 0 & 0 
				& 0 & 0	& 0 
				& \lambda_{11} & 0 & 0 
				& 0 & \lambda_{7,11,15}	& 0 
				& 0 & 0	& \lambda_{3,5,11} 
				& 0 & 0	& 0 
				& 0 & 0	& \lambda_{4,6,15} 
				& 0 & \lambda_{1,2,7} & 0 \\
		3,3,3		& 0 & 0	& \lambda_{7,10,13}
				& 0 & 0	& 0 
				& \lambda_{9,14,15} & 0	& 0 
				& 0 & 0	& 0 
				& 0 & 0	& \lambda_{7,10,13} 
				& 0 & \lambda_{9,14,15}	& 0 
				& \lambda_{8,11,12} & 0	& 0 
				& 0 & \lambda_{8,11,12} & 0 
				& 0 & 0	& \lambda_{1,\dots,15} 
	\end{bNiceArray}
	$}
\end{equation}
%\end{figure}
for scalars $\lambda_1, \dots, \lambda_{15} \in \mathbb{R}$.

Notice that $\End_{O(3)}((\mathbb{R}^{3})^{\otimes 3})$ is a $15$-dimensional vector space living inside a $729$-dimensional vector space, $\End((\mathbb{R}^{3})^{\otimes 3})$.

\subsection{$Sp(n)$}

We saw that this is similar to $O(n)$, except we replace $\delta$ by $\epsilon$ if there is an edge between two vertices that are in the same row.

%\underline{1. A Spanning Set for
%$\Hom_{Sp(2)}((\mathbb{R}^{2})^{\otimes 3}, \mathbb{R}^{2})$}
\subsubsection{A Spanning Set for
$\Hom_{Sp(2)}((\mathbb{R}^{2})^{\otimes 3}, \mathbb{R}^{2})$}

For the surjective map
\begin{equation}
	X_{3,2}^1 : B_3^1(2) \rightarrow 
		\Hom_{Sp(2)}((\mathbb{R}^{2})^{\otimes 3}, \mathbb{R}^{2})
\end{equation}
we apply Theorem \ref{spanningsetSp(n)},
noting that $l + k = 4$, which is even.

Figure \ref{Sp2matrix3,1} shows how to find a spanning set for 
$\Hom_{Sp(2)}((\mathbb{R}^{2})^{\otimes 3}, \mathbb{R}^{2})$.

\begin{figure}[ht]
	\begin{center}
\begin{tblr}{
  colspec = {X[c,h]X[c]X[2,c]},
  stretch = 0,
  rowsep = 6pt,
  hlines = {1pt},
  vlines = {1pt},
}
	{Basis Diagram $d_\beta$} 	& {Matrix Entries}	& 
	{Spanning Set Element of $\Hom_{Sp(2)}((\mathbb{R}^{2})^{\otimes 3}, \mathbb{R}^{2})$} \\
	\scalebox{0.6}{\begin{tikzpicture}
	\begin{pgfonlayer}{nodelayer}
		\node [style=none] (0) at (3.5, 1) {$1$};
		\node [style=node] (1) at (3.5, 0) {};
		\node [style=node] (2) at (3.5, -3) {};
		\node [style=node] (3) at (0.5, -3) {};
		\node [style=node] (4) at (6.5, -3) {};
		\node [style=none] (5) at (0.5, -4) {$2$};
		\node [style=none] (6) at (3.5, -4) {$3$};
		\node [style=none] (7) at (6.5, -4) {$4$};
		\node [style=none] (8) at (0, 0.5) {};
		\node [style=none] (9) at (0, -3.5) {};
		\node [style=none] (10) at (7, -3.5) {};
		\node [style=none] (11) at (7, 0.5) {};
	\end{pgfonlayer}
	\begin{pgfonlayer}{edgelayer}
		\draw [style={dash_2}] (8.center)
			 to (9.center)
			 to (10.center)
			 to (11.center)
			 to cycle;
		\draw [style=thick] (2) to (4);
		\draw [style=thick] (3) to (1);
	\end{pgfonlayer}
\end{tikzpicture}} & $(\delta_{i_1, j_1}\epsilon_{j_2,j_3})$
	& 
	\scalebox{0.75}{
	$
	\NiceMatrixOptions{code-for-first-row = \scriptstyle \color{blue},
                   	   code-for-first-col = \scriptstyle \color{blue}
	}
	\begin{bNiceArray}{*{2}{c}|*{2}{c}|*{2}{c}|*{2}{c}}[first-row,first-col]
	%	\RowStyle[cell-space-limits=3pt]{\rotate}
				& 1,1,1 	
				& 1,1,1'
				& 1,1',1 	
				& 1,1',1'
				& 1',1,1 	
				& 1',1,1'
				& 1',1',1 	
				& 1',1',1'\\
		1		
		& 0	& 1	& -1	& 0     & 0	& 0	& 0	& 0	\\
		1'		
		& 0	& 0	& 0	& 0     & 0	& 1	& -1	& 0	
	\end{bNiceArray}
	$}
	\\
	\scalebox{0.6}{\begin{tikzpicture}
	\begin{pgfonlayer}{nodelayer}
		\node [style=none] (0) at (3.5, 1) {$1$};
		\node [style=node] (1) at (3.5, 0) {};
		\node [style=node] (2) at (3.5, -3) {};
		\node [style=node] (3) at (0.5, -3) {};
		\node [style=node] (4) at (6.5, -3) {};
		\node [style=none] (5) at (0.5, -4) {$2$};
		\node [style=none] (6) at (3.5, -4) {$3$};
		\node [style=none] (7) at (6.5, -4) {$4$};
		\node [style=none] (8) at (0, 0.5) {};
		\node [style=none] (9) at (0, -3.5) {};
		\node [style=none] (10) at (7, -3.5) {};
		\node [style=none] (11) at (7, 0.5) {};
	\end{pgfonlayer}
	\begin{pgfonlayer}{edgelayer}
		\draw [style={dash_2}] (8.center)
			 to (9.center)
			 to (10.center)
			 to (11.center)
			 to cycle;
		\draw [style=thick] (2) to (1);
		\draw [style=thick, bend left=315, looseness=0.75] (4) to (3);
	\end{pgfonlayer}
\end{tikzpicture}} & $(\delta_{i_1, j_2}\epsilon_{j_1,j_3})$
	& 
	\scalebox{0.75}{
	$
	\NiceMatrixOptions{code-for-first-row = \scriptstyle \color{blue},
                   	   code-for-first-col = \scriptstyle \color{blue}
	}
	\begin{bNiceArray}{*{2}{c}|*{2}{c}|*{2}{c}|*{2}{c}}[first-row,first-col]
	%	\RowStyle[cell-space-limits=3pt]{\rotate}
				& 1,1,1 	
				& 1,1,1'
				& 1,1',1 	
				& 1,1',1'
				& 1',1,1 	
				& 1',1,1'
				& 1',1',1 	
				& 1',1',1'\\
		1		
		& 0	& 1	& 0	& 0     & -1	& 0	& 0	& 0	\\
		1'		
		& 0	& 0	& 0	& 1     & 0	& 0	& -1	& 0	
	\end{bNiceArray}
	$}
	\\
	\scalebox{0.6}{\begin{tikzpicture}
	\begin{pgfonlayer}{nodelayer}
		\node [style=none] (0) at (3.5, 1) {$1$};
		\node [style=node] (1) at (3.5, 0) {};
		\node [style=node] (2) at (3.5, -3) {};
		\node [style=node] (3) at (0.5, -3) {};
		\node [style=node] (4) at (6.5, -3) {};
		\node [style=none] (5) at (0.5, -4) {$2$};
		\node [style=none] (6) at (3.5, -4) {$3$};
		\node [style=none] (7) at (6.5, -4) {$4$};
		\node [style=none] (8) at (0, 0.5) {};
		\node [style=none] (9) at (0, -3.5) {};
		\node [style=none] (10) at (7, -3.5) {};
		\node [style=none] (11) at (7, 0.5) {};
	\end{pgfonlayer}
	\begin{pgfonlayer}{edgelayer}
		\draw [style={dash_2}] (8.center)
			 to (9.center)
			 to (10.center)
			 to (11.center)
			 to cycle;
		\draw [style=thick] (1) to (4);
		\draw [style=thick] (2) to (3);
	\end{pgfonlayer}
\end{tikzpicture}} & $(\delta_{i_1, j_3}\epsilon_{j_1,j_2})$
	& 
	\scalebox{0.75}{
	$
	\NiceMatrixOptions{code-for-first-row = \scriptstyle \color{blue},
                   	   code-for-first-col = \scriptstyle \color{blue}
	}
	\begin{bNiceArray}{*{2}{c}|*{2}{c}|*{2}{c}|*{2}{c}}[first-row,first-col]
	%	\RowStyle[cell-space-limits=3pt]{\rotate}
				& 1,1,1 	
				& 1,1,1'
				& 1,1',1 	
				& 1,1',1'
				& 1',1,1 	
				& 1',1,1'
				& 1',1',1 	
				& 1',1',1'\\
		1		
		& 0	& 0	& 1	& 0     & -1	& 0	& 0	& 0	\\
		1'		
		& 0	& 0	& 0	& 1     & 0	& -1	& 0 	& 0	
	\end{bNiceArray}
	$}
	\\
\end{tblr}
		\caption{The images under $X_{3,2}^1$ of the basis diagrams of $B_3^1(2)$ make up a spanning set for $\Hom_{Sp(2)}((\mathbb{R}^{2})^{\otimes 3}, \mathbb{R}^{2})$.}
			%A table showing the basis diagrams of $B_3^1(2)$ whose images under $X_{3,2}^1$ make up a spanning set for $\Hom_{Sp(2)}((\mathbb{R}^{2})^{\otimes 3}, \mathbb{R}^{2})$.}
  	\label{Sp2matrix3,1}
	\end{center}
\end{figure}

This means that any element of 
$\Hom_{Sp(2)}((\mathbb{R}^{2})^{\otimes 3}, \mathbb{R}^{2})$,
in the basis of matrix units of 
$\Hom((\mathbb{R}^{2})^{\otimes 3}, \mathbb{R}^{2})$,
is of the form
%\begin{figure}[ht]
\begin{equation}
	\NiceMatrixOptions{code-for-first-row = \scriptstyle \color{blue},
                   	   code-for-first-col = \scriptstyle \color{blue}
	}
	\renewcommand{\arraystretch}{1.5}
	\begin{bNiceArray}{*{2}{c}|*{2}{c}|*{2}{c}|*{2}{c}}[first-row,first-col]
%		\RowStyle[cell-space-limits=3pt]{\rotate}
				& 1,1,1 	
				& 1,1,1'
				& 1,1',1 	
				& 1,1',1'
				& 1',1,1 	
				& 1',1,1'
				& 1',1',1 	
				& 1',1',1'\\
		1		
		& 0	& \lambda_{1} + \lambda_{2}	& - \lambda_1 + \lambda_3
		& 0     & - \lambda_{2} - \lambda_{3}	& 0	& 0	& 0	\\
		1'		
		& 0	& 0	& 0	& \lambda_{2} + \lambda_{3}    & 0	& \lambda_{1} - \lambda_{3}	& - \lambda_{1} - \lambda_{2} 	& 0	
	\end{bNiceArray}
\end{equation}
%\end{figure}
for scalars $\lambda_1, \lambda_2, \lambda_3 \in \mathbb{R}$.

\subsection{$SO(n)$}

%\underline{1. A Spanning Set for
%$\End_{SO(3)}((\mathbb{R}^{3})^{\otimes 3})$}
\subsubsection{A Spanning Set for
$\End_{SO(3)}((\mathbb{R}^{3})^{\otimes 3})$}

We apply Theorem \ref{spanningsetSO(n)}.

As $n \leq l + k$, and $n$ is odd and $l + k$ is even, we see that 
$\End_{SO(3)}((\mathbb{R}^{3})^{\otimes 3}) =
\End_{O(3)}((\mathbb{R}^{3})^{\otimes 3})$
and so any element of 
$\End_{SO(3)}((\mathbb{R}^{3})^{\otimes 3})$, in the basis of matrix units of
$\End((\mathbb{R}^{3})^{\otimes 3})$,
is of the form (\ref{EndO3R3tensor3}).

%\underline{2. A Spanning Set for
%$\Hom_{SO(2)}((\mathbb{R}^{2})^{\otimes 3}, \mathbb{R}^{2})$}
\subsubsection{A Spanning Set for
$\Hom_{SO(2)}((\mathbb{R}^{2})^{\otimes 3}, \mathbb{R}^{2})$}

Again, we apply Theorem \ref{spanningsetSO(n)}.

As $n \leq l + k$, and $n$ is even and $l + k$ is even, we see that 
three $(3,1)$--Brauer diagrams and six $(1+3)\backslash 2$--diagrams make up a basis of $D_3^1(2)$. 
Their images under 
\begin{equation}
	\Psi_{3,2}^1 : D_3^1(2) \rightarrow 
		\Hom_{SO(2)}((\mathbb{R}^{2})^{\otimes 3}, \mathbb{R}^{2})
\end{equation}
forms a spanning set of 
$\Hom_{SO(2)}((\mathbb{R}^{2})^{\otimes 3}, \mathbb{R}^{2})$.

Calculating the images of the three $(3,1)$--Brauer diagrams is the same as for the $O(n)$ case.
Figure \ref{SO2Brauermatrix3,1} shows how to find the images of the six
$(1+3)\backslash 2$--diagrams.

\subsection{Local Symmetry Example}

As an example of the result given in Section~\ref{EquivLocal}, suppose that we want to find a spanning set for
\begin{equation} \label{localsymmex}
	\Hom_{SO(3) \times SO(3)}((\mathbb{R}^{3})^{\otimes {3}} \boxtimes \mathbb{R}^{3}, 
(\mathbb{R}^{3})^{\otimes {3}} \boxtimes (\mathbb{R}^{3})^{\otimes {2}})
\end{equation}
By (\ref{tensorgenericHomspace}),
we know that (\ref{localsymmex}) is isomorphic to
\begin{equation} 
	\End_{SO(3)}((\mathbb{R}^{3})^{\otimes {3}}) \otimes
	\Hom_{SO(3)}(\mathbb{R}^{3}, (\mathbb{R}^{3})^{\otimes {2}})
\end{equation}
Hence, to find a spanning set, all we need to do is find the images of the basis elements of
$D_3^3(3) \otimes D_1^2(3)$ under $\Psi_{3,3}^3 \otimes \Psi_{1,3}^2$ and take the Kronecker product of the resulting matrices.

To save space, we will only show what the basis elements of 
$D_3^3(3) \otimes D_1^2(3)$ are.
By Theorem \ref{spanningsetSO(n)}, the basis elements are
\begin{center}
	\scalebox{0.45}{\input{brauer33grood12.tikz}}
\end{center}
where we have used a red demarcation line to separate the vertices of the respective diagrams. Note that no edge can cross this red line.

\begin{figure}[ht]
	\begin{center}
\begin{tblr}{
  %colspec = {X[c,h]X[c,m]X[c,m]},
  colspec = {X[c,h]X[c]X[2,c]},
  stretch = 0,
  rowsep = 6pt,
  hlines = {1pt},
  vlines = {1pt},
}
	{Basis Diagram $d_\alpha$} 	& {Matrix Entries}	& 
	{Spanning Set Element of $\Hom_{SO(2)}((\mathbb{R}^{2})^{\otimes 3}, \mathbb{R}^{2})$} \\
	\scalebox{0.6}{\begin{tikzpicture}
	\begin{pgfonlayer}{nodelayer}
		\node [style=none] (0) at (5, 0) {$1$};
		\node [style=none] (1) at (2, -5) {$2$};
		\node [style=none] (2) at (5, -5) {$3$};
		\node [style=none] (3) at (8, -5) {$4$};
		\node [style=node] (10) at (5, -1) {};
		\node [style=node] (14) at (2, -4) {};
		\node [style=node] (15) at (5, -4) {};
		\node [style=node] (16) at (8, -4) {};
		\node [style=none] (20) at (1.5, -0.5) {};
		\node [style=none] (21) at (1.5, -4.5) {};
		\node [style=none] (22) at (8.5, -4.5) {};
		\node [style=none] (23) at (8.5, -0.5) {};
	\end{pgfonlayer}
	\begin{pgfonlayer}{edgelayer}
		\draw [style={dash_3}] (21.center)
			 to (22.center)
			 to (23.center)
			 to (20.center)
			 to cycle;
		\draw [style=thick] (14) to (10);
	\end{pgfonlayer}
\end{tikzpicture}} & 
	$(
		\chi
			\left(\begin{smallmatrix} 
				1 & 2 \\
				j_2 & j_3 
			\end{smallmatrix}\right)
	\delta_{i_1,j_1})$
	& 
	\scalebox{0.75}{
	$
	\NiceMatrixOptions{code-for-first-row = \scriptstyle \color{blue},
                   	   code-for-first-col = \scriptstyle \color{blue}
	}
	\begin{bNiceArray}{*{2}{c}|*{2}{c}|*{2}{c}|*{2}{c}}[first-row,first-col]
%		\RowStyle[cell-space-limits=3pt]{\rotate}
				& 1,1,1 	
				& 1,1,2
				& 1,2,1 	
				& 1,2,2
				& 2,1,1 	
				& 2,1,2
				& 2,2,1 	
				& 2,2,2\\
		1		
		& 0	& 1	& -1	& 0     & 0	& 0	& 0	& 0	\\
		2		
		& 0	& 0	& 0	& 0     & 0	& 1	& -1 	& 0	
	\end{bNiceArray}
	$}
	\\
	\scalebox{0.6}{\begin{tikzpicture}
	\begin{pgfonlayer}{nodelayer}
		\node [style=none] (0) at (5, 0) {$1$};
		\node [style=none] (1) at (2, -5) {$2$};
		\node [style=none] (2) at (5, -5) {$3$};
		\node [style=none] (3) at (8, -5) {$4$};
		\node [style=node] (4) at (5, -1) {};
		\node [style=node] (5) at (2, -4) {};
		\node [style=node] (6) at (5, -4) {};
		\node [style=node] (7) at (8, -4) {};
		\node [style=none] (8) at (1.5, -0.5) {};
		\node [style=none] (9) at (1.5, -4.5) {};
		\node [style=none] (10) at (8.5, -4.5) {};
		\node [style=none] (11) at (8.5, -0.5) {};
	\end{pgfonlayer}
	\begin{pgfonlayer}{edgelayer}
		\draw [style={dash_3}] (8.center)
			 to (9.center)
			 to (10.center)
			 to (11.center)
			 to cycle;
		\draw [style=thick] (4) to (6);
	\end{pgfonlayer}
\end{tikzpicture}} & 
	$(
		\chi
			\left(\begin{smallmatrix} 
				1 & 2 \\
				j_1 & j_3 
			\end{smallmatrix}\right)
	\delta_{i_1,j_2})$
	& 
	\scalebox{0.75}{
	$
	\NiceMatrixOptions{code-for-first-row = \scriptstyle \color{blue},
                   	   code-for-first-col = \scriptstyle \color{blue}
	}
	\begin{bNiceArray}{*{2}{c}|*{2}{c}|*{2}{c}|*{2}{c}}[first-row,first-col]
%		\RowStyle[cell-space-limits=3pt]{\rotate}
				& 1,1,1 	
				& 1,1,2
				& 1,2,1 	
				& 1,2,2
				& 2,1,1 	
				& 2,1,2
				& 2,2,1 	
				& 2,2,2\\
		1		
		& 0	& 1	& 0	& 0     & -1	& 0	& 0	& 0	\\
		2		
		& 0	& 0	& 0	& 1     & 0	& 0	& -1 	& 0	
	\end{bNiceArray}
	$}
	\\
	\scalebox{0.6}{\begin{tikzpicture}
	\begin{pgfonlayer}{nodelayer}
		\node [style=none] (0) at (5, 0) {$1$};
		\node [style=none] (1) at (2, -5) {$2$};
		\node [style=none] (2) at (5, -5) {$3$};
		\node [style=none] (3) at (8, -5) {$4$};
		\node [style=node] (4) at (5, -1) {};
		\node [style=node] (5) at (2, -4) {};
		\node [style=node] (6) at (5, -4) {};
		\node [style=node] (7) at (8, -4) {};
		\node [style=none] (8) at (1.5, -0.5) {};
		\node [style=none] (9) at (1.5, -4.5) {};
		\node [style=none] (10) at (8.5, -4.5) {};
		\node [style=none] (11) at (8.5, -0.5) {};
	\end{pgfonlayer}
	\begin{pgfonlayer}{edgelayer}
		\draw [style={dash_3}] (8.center)
			 to (9.center)
			 to (10.center)
			 to (11.center)
			 to cycle;
		\draw [style=thick] (4) to (7);
	\end{pgfonlayer}
\end{tikzpicture}} & 
	$(
		\chi
			\left(\begin{smallmatrix} 
				1 & 2 \\
				j_1 & j_2 
			\end{smallmatrix}\right)
	\delta_{i_1,j_3})$
	& 
	\scalebox{0.75}{
	$
	\NiceMatrixOptions{code-for-first-row = \scriptstyle \color{blue},
                   	   code-for-first-col = \scriptstyle \color{blue}
	}
	\begin{bNiceArray}{*{2}{c}|*{2}{c}|*{2}{c}|*{2}{c}}[first-row,first-col]
%		\RowStyle[cell-space-limits=3pt]{\rotate}
				& 1,1,1 	
				& 1,1,2
				& 1,2,1 	
				& 1,2,2
				& 2,1,1 	
				& 2,1,2
				& 2,2,1 	
				& 2,2,2\\
		1		
		& 0	& 0	& 1	& 0     & -1	& 0	& 0	& 0	\\
		2		
		& 0	& 0	& 0	& 1     & 0	& -1	& 0 	& 0	
	\end{bNiceArray}
	$}
	\\
	\scalebox{0.6}{\begin{tikzpicture}
	\begin{pgfonlayer}{nodelayer}
		\node [style=none] (0) at (5, 0) {$1$};
		\node [style=none] (1) at (2, -5) {$2$};
		\node [style=none] (2) at (5, -5) {$3$};
		\node [style=none] (3) at (8, -5) {$4$};
		\node [style=node] (4) at (5, -1) {};
		\node [style=node] (5) at (2, -4) {};
		\node [style=node] (6) at (5, -4) {};
		\node [style=node] (7) at (8, -4) {};
		\node [style=none] (8) at (1.5, -0.5) {};
		\node [style=none] (9) at (1.5, -4.5) {};
		\node [style=none] (10) at (8.5, -4.5) {};
		\node [style=none] (11) at (8.5, -0.5) {};
	\end{pgfonlayer}
	\begin{pgfonlayer}{edgelayer}
		\draw [style={dash_3}] (9.center)
			 to (10.center)
			 to (11.center)
			 to (8.center)
			 to cycle;
		\draw [style=thick] (5) to (6);
	\end{pgfonlayer}
\end{tikzpicture}} & 
	$(
		\chi
			\left(\begin{smallmatrix} 
				1 & 2 \\
				i_1 & j_3 
			\end{smallmatrix}\right)
	\delta_{j_1,j_2})$
	& 
	\scalebox{0.75}{
	$
	\NiceMatrixOptions{code-for-first-row = \scriptstyle \color{blue},
                   	   code-for-first-col = \scriptstyle \color{blue}
	}
	\begin{bNiceArray}{*{2}{c}|*{2}{c}|*{2}{c}|*{2}{c}}[first-row,first-col]
%		\RowStyle[cell-space-limits=3pt]{\rotate}
				& 1,1,1 	
				& 1,1,2
				& 1,2,1 	
				& 1,2,2
				& 2,1,1 	
				& 2,1,2
				& 2,2,1 	
				& 2,2,2\\
		1		
		& 0	& 1	& 0	& 0     & 0	& 0	& 0	& 1	\\
		2		
		& -1	& 0	& 0	& 0     & 0	& 0	& -1 	& 0	
	\end{bNiceArray}
	$}
	\\
	\scalebox{0.6}{\begin{tikzpicture}
	\begin{pgfonlayer}{nodelayer}
		\node [style=none] (0) at (5, 0) {$1$};
		\node [style=none] (1) at (2, -5) {$2$};
		\node [style=none] (2) at (5, -5) {$3$};
		\node [style=none] (3) at (8, -5) {$4$};
		\node [style=node] (4) at (5, -1) {};
		\node [style=node] (5) at (2, -4) {};
		\node [style=node] (6) at (5, -4) {};
		\node [style=node] (7) at (8, -4) {};
		\node [style=none] (8) at (1.5, -0.5) {};
		\node [style=none] (9) at (1.5, -4.5) {};
		\node [style=none] (10) at (8.5, -4.5) {};
		\node [style=none] (11) at (8.5, -0.5) {};
	\end{pgfonlayer}
	\begin{pgfonlayer}{edgelayer}
		\draw [style={dash_3}] (9.center)
			 to (10.center)
			 to (11.center)
			 to (8.center)
			 to cycle;
		\draw [style=thick, bend left=315, looseness=0.75] (7) to (5);
	\end{pgfonlayer}
\end{tikzpicture}} & 
	$(
		\chi
			\left(\begin{smallmatrix} 
				1 & 2 \\
				i_1 & j_2 
			\end{smallmatrix}\right)
	\delta_{j_1,j_3})$
	& 
	\scalebox{0.75}{
	$
	\NiceMatrixOptions{code-for-first-row = \scriptstyle \color{blue},
                   	   code-for-first-col = \scriptstyle \color{blue}
	}
	\begin{bNiceArray}{*{2}{c}|*{2}{c}|*{2}{c}|*{2}{c}}[first-row,first-col]
%		\RowStyle[cell-space-limits=3pt]{\rotate}
				& 1,1,1 	
				& 1,1,2
				& 1,2,1 	
				& 1,2,2
				& 2,1,1 	
				& 2,1,2
				& 2,2,1 	
				& 2,2,2\\
		1		
		& 0	& 0	& 1	& 0     & 0	& 0	& 0	& 1	\\
		2		
		& -1	& 0	& 0	& 0     & 0	& -1	& 0 	& 0	
	\end{bNiceArray}
	$}
	\\
	\scalebox{0.6}{\begin{tikzpicture}
	\begin{pgfonlayer}{nodelayer}
		\node [style=none] (0) at (5, 0) {$1$};
		\node [style=none] (1) at (2, -5) {$2$};
		\node [style=none] (2) at (5, -5) {$3$};
		\node [style=none] (3) at (8, -5) {$4$};
		\node [style=node] (4) at (5, -1) {};
		\node [style=node] (5) at (2, -4) {};
		\node [style=node] (6) at (5, -4) {};
		\node [style=node] (7) at (8, -4) {};
		\node [style=none] (8) at (1.5, -0.5) {};
		\node [style=none] (9) at (1.5, -4.5) {};
		\node [style=none] (10) at (8.5, -4.5) {};
		\node [style=none] (11) at (8.5, -0.5) {};
	\end{pgfonlayer}
	\begin{pgfonlayer}{edgelayer}
		\draw [style={dash_3}] (9.center)
			 to (10.center)
			 to (11.center)
			 to (8.center)
			 to cycle;
		\draw [style=thick] (6) to (7);
	\end{pgfonlayer}
\end{tikzpicture}} & 
	$(
		\chi
			\left(\begin{smallmatrix} 
				1 & 2 \\
				i_1 & j_1 
			\end{smallmatrix}\right)
	\delta_{j_2,j_3})$
	& 
	\scalebox{0.75}{
	$
	\NiceMatrixOptions{code-for-first-row = \scriptstyle \color{blue},
                   	   code-for-first-col = \scriptstyle \color{blue}
	}
	\begin{bNiceArray}{*{2}{c}|*{2}{c}|*{2}{c}|*{2}{c}}[first-row,first-col]
%		\RowStyle[cell-space-limits=3pt]{\rotate}
				& 1,1,1 	
				& 1,1,2
				& 1,2,1 	
				& 1,2,2
				& 2,1,1 	
				& 2,1,2
				& 2,2,1 	
				& 2,2,2\\
		1		
		& 0	& 0	& 0	& 0     & 1	& 0	& 0	& 1	\\
		2		
		& -1	& 0	& 0	& -1     & 0	& 0	& 0 	& 0	
	\end{bNiceArray}
	$}
\end{tblr}
		\caption{The images under $\Psi_{3,2}^1$ of the six $(1+3) \backslash 2$--diagrams in $D_3^1(2)$, together with the images under $\Psi_{3,2}^1$ of the three $(3,1)$--Brauer diagrams in $D_3^1(2)$ (not shown), make up a spanning set for $\Hom_{SO(2)}((\mathbb{R}^{2})^{\otimes 3}, \mathbb{R}^{2})$.}
	%\caption{The images of the six $4 \backslash 2$--diagrams in $D_3^1(2)$ under $\Psi_{3,2}^1$ form part of a spanning set for $\Hom_{SO(2)}((\mathbb{R}^{2})^{\otimes 3}, \mathbb{R}^{2})$.}
			%A table showing the images of the six $4 \backslash 2$--diagrams in $D_3^1(2)$ whose images under $\Psi_{3,2}^1$ form part of a spanning set for
		%$\Hom_{SO(2)}((\mathbb{R}^{2})^{\otimes 3}, \mathbb{R}^{2})$.}
  	\label{SO2Brauermatrix3,1}
	\end{center}
\end{figure}

\end{document}